\documentclass[12pt]{article}

\usepackage{amsmath,amsfonts,bm}

\def\eqref#1{equation~\ref{#1}}
\def\1{\bm{1}}

\DeclareMathAlphabet{\mathsfit}{\encodingdefault}{\sfdefault}{m}{sl}
\SetMathAlphabet{\mathsfit}{bold}{\encodingdefault}{\sfdefault}{bx}{n}

\usepackage[left=2.5cm,
right=2.5cm,
top=2.3cm,
bottom=2.3cm,
headheight=20pt,
headsep=10pt,
footskip=25pt,
letterpaper]{geometry}
\usepackage{times}
\usepackage{helvet}
\usepackage{courier}
\usepackage[hyphens]{url}
\usepackage{graphicx}
\usepackage[numbers,sort&compress]{natbib}
\usepackage{caption}
\usepackage{algorithm}
\usepackage{algorithmic}
\usepackage{booktabs}

\usepackage[most]{tcolorbox}
\usepackage{fancyvrb}
\usepackage{enumitem}
\usepackage{fvextra}
\usepackage{amssymb}
\usepackage{multirow}
\DefineVerbatimEnvironment{Verbatim}{Verbatim}{breaklines=true, breakanywhere=true}

\usepackage{tabularx, array}
\usepackage{ragged2e}
\usepackage{amsmath}
\usepackage{subcaption}
\newcolumntype{Y}{>{\centering\arraybackslash}X}

\graphicspath{{fig/}}

\usepackage{makecell}

\usepackage{newfloat}
\usepackage{listings}
\DeclareCaptionStyle{ruled}{labelfont=normalfont,labelsep=colon,strut=off}
\floatstyle{ruled}
\newfloat{listing}{tb}{lst}{}
\floatname{listing}{Listing}

\usepackage{xcolor}
\usepackage[utf8]{inputenc}
\usepackage[T1]{fontenc}
\usepackage[english]{babel}
\usepackage{amsmath,amsfonts,amssymb,thmtools,amsthm}
\usepackage{hyperref}
\usepackage{fancyhdr}
\usepackage{stfloats}
\usepackage{wrapfig}
\usepackage{epstopdf}
\usepackage{cleveref}
\usepackage{subfloat}
\usepackage{titlesec}
\usepackage{etoolbox}
\usepackage{setspace}
\usepackage{changepage}
\usepackage{svg}
\usepackage[percent]{overpic}
\usepackage[colorinlistoftodos, shadow, color=blue!30!white, disable]{todonotes}
\usepackage{xpatch}
\usepackage{siunitx}

\renewcommand{\sfdefault}{phv}

\renewcommand{\headrulewidth}{0pt}
\renewcommand{\footrulewidth}{0pt}
\fancypagestyle{first}{\fancyfoot[R]{\small\thepage}}

\setlist[itemize]{leftmargin=1em,itemsep=0ex,topsep=0ex}
\titlespacing*{\paragraph}{0pt}{0ex plus .1ex}{1ex}
\titlespacing*{\section}{0ex}{2.3ex plus .3ex minus .0ex}{.6ex plus .3ex minus .2ex}
\titlespacing*{\subsection}{0ex}{1.5ex plus .3ex minus .5ex}{.4ex plus .2ex minus .1ex}
\titlespacing*{\subsubsection}{0ex}{1.2ex plus .3ex minus .3ex}{.3ex plus .2ex minus .2ex}

\xapptocmd\normalsize{%
\abovedisplayskip=.8em plus .2em minus .2em
\belowdisplayskip=.6em plus .1em minus .1em
\abovedisplayshortskip=.8em plus .2em minus .2em
\belowdisplayshortskip=.6em plus .1em minus .1em
}{}{}

\renewcommand{\cite}[1]{\citep{#1}}

\definecolor{mydarkblue}{rgb}{0.0,0.15,0.7}
\hypersetup{%
colorlinks=true,
linkcolor=mydarkblue,
citecolor=mydarkblue,
filecolor=mydarkblue,
urlcolor=mydarkblue}

\makeatletter

  \renewcommand{\maketitle}{%
    \begingroup
      {\centering\Large\@title\par}%
      \vskip 0.2em
      \centering
      \begin{tabular}[t]{@{}c@{}}\strut\@author\strut\end{tabular}%
      \vskip 0.3in minus 0.1in
    \endgroup
  }
\makeatother

\newcolumntype{L}{>{\RaggedRight\arraybackslash}X}

\usepackage{pgfplots}
\usepackage{fontawesome5}
\pgfplotsset{width=12cm, height=8cm, compat=1.18}
\usepackage{tikz}
\usetikzlibrary{calc}
\usetikzlibrary{positioning, shapes, fit, backgrounds, decorations.pathreplacing}
\usepackage{colortbl}
\usepackage{longtable}
\usepackage{markdown}
\usepackage{pifont}
\usepackage{microtype}
\usepackage{nicefrac}
\usepackage{float}
\usepackage{xspace}
\usepackage{tablefootnote}

\definecolor{titlecolor}{RGB}{255,170,100}
\definecolor{abstractcolor}{RGB}{255,119,51}
\definecolor{introductioncolor}{RGB}{188,56,0}
\definecolor{backgroundcolor}{RGB}{255,153,102}
\definecolor{evaluationcolor}{RGB}{216,84,0}
\definecolor{conclusioncolor}{RGB}{255,200,150}

\newcommand{\model}{$\texttt{XtraGPT}$~}

\def\huggingface{\raisebox{-1.5pt}{\includegraphics[height=1.05em]{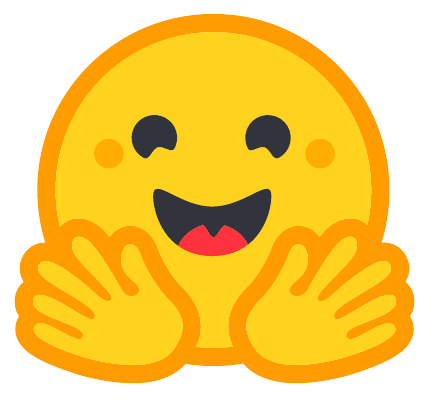}}}
\def\github{\raisebox{-1.5pt}{\includegraphics[height=1.05em]{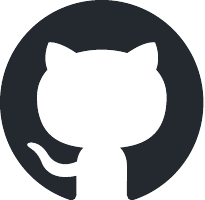}}}

\tcbset{
    aibox/.style={
        width=\textwidth,
        top=10pt,
        colback=white,
        colframe=black,
        colbacktitle=black,
        enhanced,
        center,
        attach boxed title to top left={yshift=-0.12in,xshift=0.15in},
        boxed title style={boxrule=0pt,colframe=white},
    }
}
\newtcolorbox{AIbox}[2][]{aibox,title=#2,#1}

\title{\protect\includegraphics[width=0.5cm]{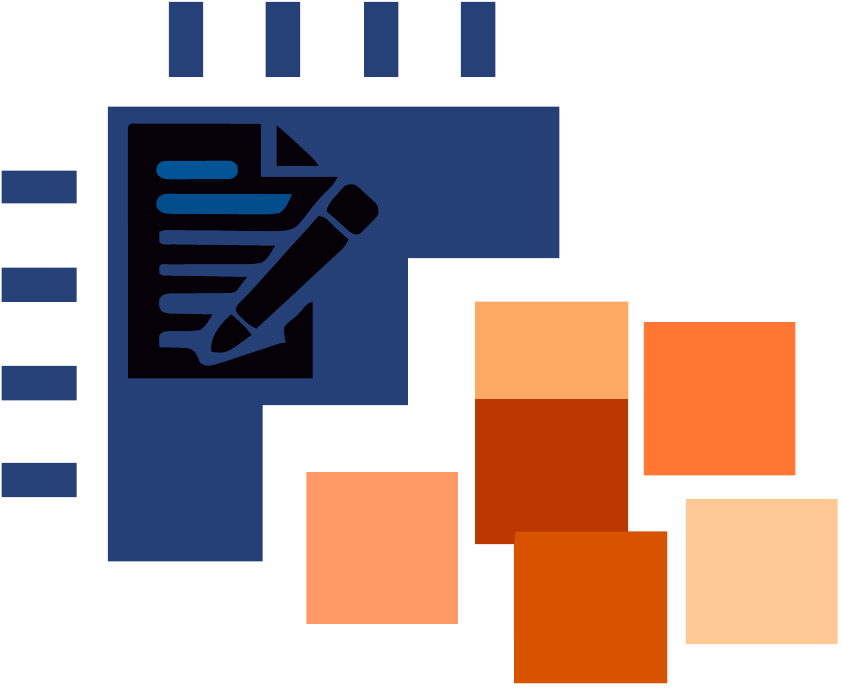} \model: Context-Aware and Controllable \\ Academic Paper Revision via Human-AI Collaboration}

\date{}
\author{
\begin{tabular}{cccc}
Nuo Chen$^{1}$ & Andre Huikai Lin$^{1}$ & Jiaying Wu$^{1}$ & Junyi Hou$^{1}$ \\
Zining Zhang$^{1}$ & Qian Wang$^{1}$ & Xidong Wang$^{2}$ & Bingsheng He$^{1}$
\end{tabular}
\\[6pt]
$^1$National University of Singapore \quad $^2$The Chinese University of Hong Kong, Shenzhen
\\
\github \ \textbf{Code}: \url{https://github.com/Xtra-Computing/XtraGPT}
\\
\huggingface\ \textbf{\protect\includegraphics[width=1.8cm]{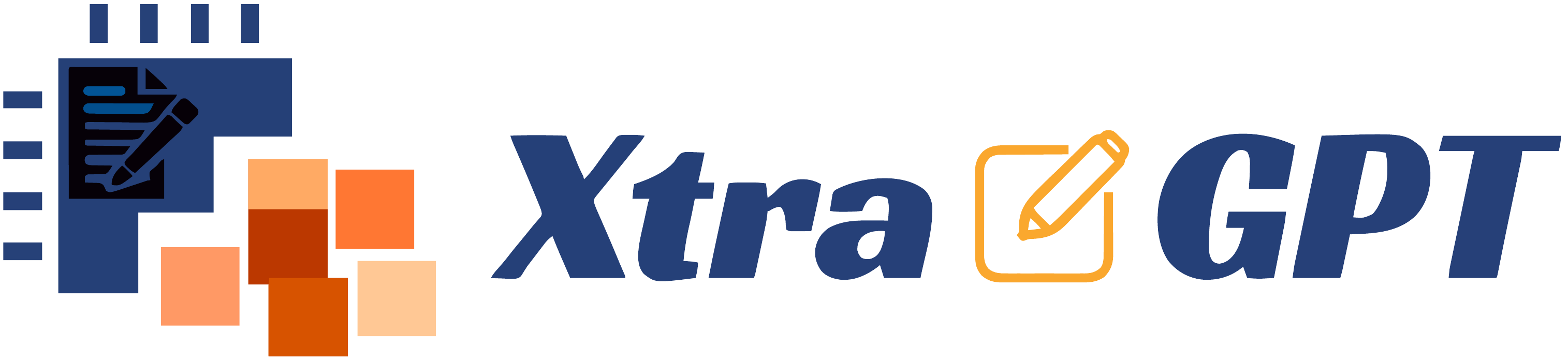} Collection}: \href{https://huggingface.co/collections/Xtra-Computing/xtragpt}{\nolinkurl{Xtra-Computing/xtragpt}} \quad  \faTools \ \href{https://www.npmjs.com/package/xtragpt-paper-revision-skill}{\nolinkurl{xtragpt-paper-revision-skill}}
}

\begin{document}
\pagestyle{fancy}

\maketitle
\thispagestyle{first}

\vspace{-3mm}
\begin{figure}[h]
    \centering
    \vspace{-4.5mm}
    \includegraphics[width=0.98\linewidth]{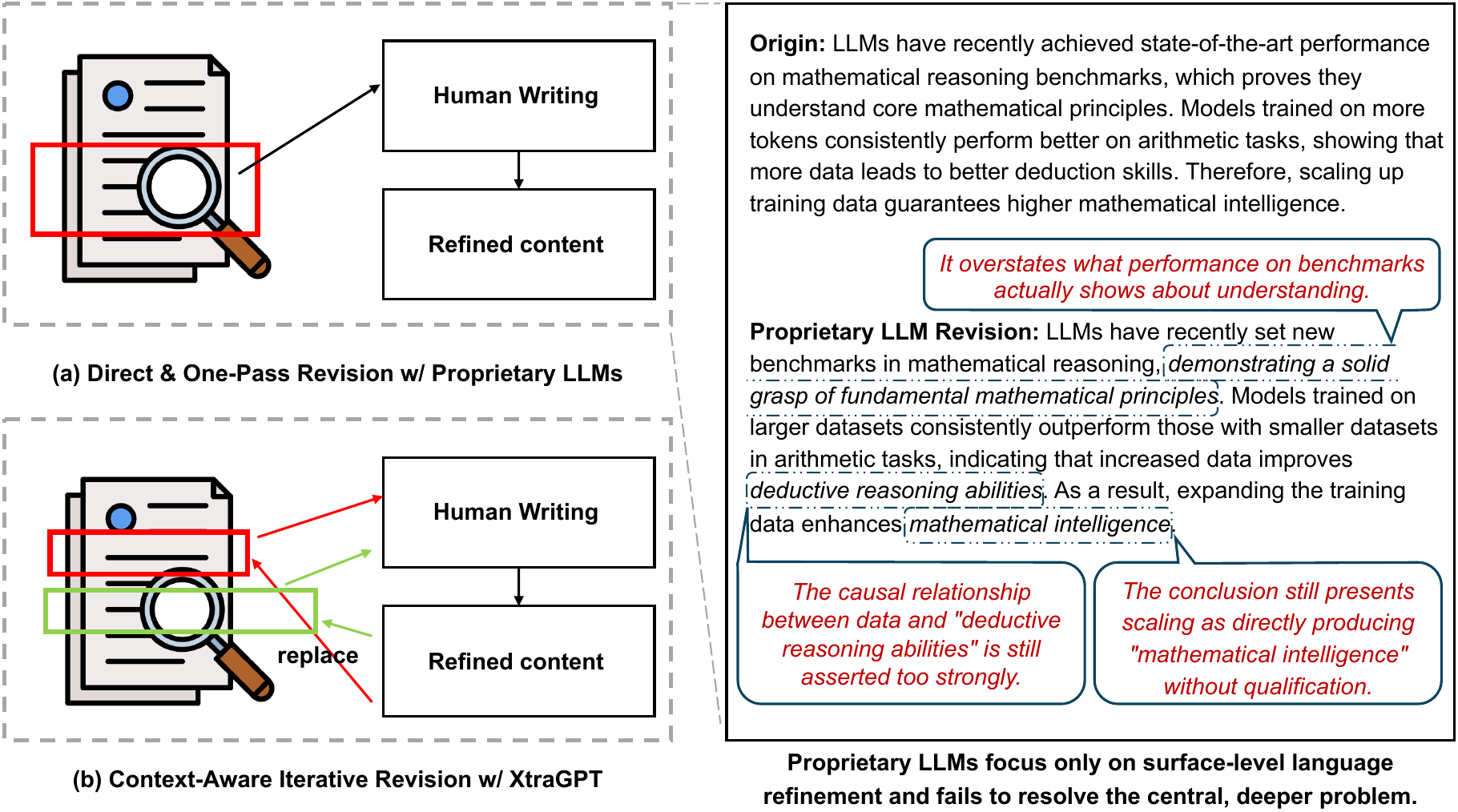}
    \caption{(Left) Overview of the academic paper revision workflow comparing proprietary LLMs and our method. (Right) An example of a poor revision generated by a proprietary LLM. A detailed case study of our model \texttt{XtraGPT} is provided in Table~\ref{tab:casestudy}.}
    \label{fig:intro}
\end{figure}
\vspace{-3mm}
\begin{abstract}

Despite the growing adoption of large language models (LLMs) in academic workflows, their capabilities remain limited in supporting high-quality scientific writing. Most existing systems are designed for general-purpose scientific text generation and fail to meet the sophisticated demands of research communication beyond surface-level polishing, for example, maintaining conceptual coherence across sections. Furthermore, academic writing is inherently iterative and revision-driven, a process that is not well supported by direct prompting-based paradigms. To address these scenarios, we propose a human-AI collaboration framework for academic paper revision, centered on criteria-guided intent alignment and context-aware modeling. To validate the framework, we curate a dataset of 7,000 research papers from top-tier venues, annotated with 140,000 instruction--response pairs that reflect realistic, section-level scientific revisions.
We instantiate the framework in \texttt{XtraGPT},  the first suite of open-source LLMs (1.5B to 14B parameters) specifically fine-tuned for context-aware academic paper revision.
 Extensive experiments show that \texttt{XtraGPT} significantly outperforms same-scale baselines and rivals the quality of proprietary counterparts. Both automated preference assessments and human evaluations confirm the effectiveness of \texttt{XtraGPT} in improving scientific drafts.
\end{abstract}

\section{Introduction}
Large language models (LLMs) are increasingly embedded in cognitively demanding workflows, particularly in scientific domains such as hypothesis generation \citep{zhou2025hypothesis}, proposal writing \citep{gottweis2025towards}, and literature review \citep{asai2024openscholar}. While some applications focus on generating scientific content from scratch \citep{weng2024cycleresearcher,yamada2025aiscientistv2,wang2025llmtimesmapreducev2entropydrivenconvolutionaltesttime} (ethical concerns are discussed in Appendix \ref{sec:impact}), researchers increasingly leverage LLMs as assistants for refining their academic paper drafts. In this setting, users request specific improvements (such as enhancing clarity or strengthening the motivation), and the model suggests revisions grounded in the original draft, as illustrated in Figure \ref{fig:intro} (a). This human-AI collaborative revision process \citep{khalifa2024using,naddaf2025how} helps preserve scientific originality while improving the clarity, coherence, and overall presentation of research ideas.

The prevailing mode of LLM-assisted paper revision involves simply prompting models such as GPT-4o \citep{hurst2024gpt} through web interfaces. This approach is effective for surface-level editing, but faces two key limitations in the context of scientific writing. First, \textbf{general-purpose LLMs often lack explicit understanding of the deeper structure and argumentative rigor required for academic papers}. As shown in Figure~\ref{fig:intro}, when prompted to revise an unclear motivation paragraph, GPT-4o improved the fluency of the text but failed to address the missing rationale, leaving the revised version equally unpersuasive or, at times, overly belabored in its points. Second, \textbf{scientific writing is inherently an iterative process} \citep{montgomery2017chicago,schimel2012writing} rather than generating from scratch. Authors revise their work over multiple rounds of drafting, incorporating feedback and improving content in a context-sensitive manner. Current LLM workflows do not naturally support this iterative nature \citep{wang2026liveclin}. They treat each prompt in isolation and lack mechanisms for tracking changes or maintaining context across revision cycles. This limits their effectiveness in real-world scientific authoring scenarios.

Human-AI Collaborative (HAC) paper revision demands an instruction-driven, fine-grained process that aligns closely with established norms of scientific writing. These norms are often formalized in detailed guidelines issued by top-tier conferences, such as the ICLR reviewer guide \citep{iclr2025reviewerguide}. While recent studies have explored related tasks including end-to-end article generation \citep{weng2024cycleresearcher}, idea generation \citep{ghafarollahi2024sciagents}, automated paper evaluation \citep{zhu2025deepreview}, and grammar correction \citep{TeXGPT}, they do not explicitly model the rationale-driven, feedback-centric nature of scientific revision. Specifically, they lack controllability in three key aspects: (1) following in-context examples, (2) adhering to user instructions, and (3) generating outputs that conform to explicit writing criteria. As a result, their generated content often suffers in clarity, coherence, or alignment with scholarly standards, as summarized in Table~\ref{tab:comparison}.

To make LLMs more effective and accessible for scientific writing support, we propose a framework in which human authors retain creative control by generating ideas and drafting content, while LLMs act as targeted assistants that provide context-aware revisions. Our goal is to assist authors in producing improved writing while keeping the additional overhead of the writing process minimal. In Section~\ref{sec:framework}, we detail the high-level principles and methodological design, emphasizing post-training for enhanced in-context learning (ICL) to enable controllable, iterative revisions via human-AI collaboration. On the data collection protocol, to simulate real-world scientific revision needs, each revision is guided by one of 20 section-level criteria informed by authoritative writing guides~\citep{widom2006tips} and expert revisions contributed by experienced AI researchers (see Table~\ref{tab:commentsdetails}). These pairs cover a diverse range of instruction-driven, section-specific edits grounded in realistic revision scenarios. Furthermore, we provide an evaluation framework to measure the controllability and effectiveness using length-controlled LLM-as-a-Judge, allowing future developments to be benchmarked and justified.

\begin{small}
    \begin{table*}[t]
    \centering
        \caption{Comparison of existing full-paper AI generation approaches across key dimensions: common quality issues, use of In-Context Learning (ICL), specific writing tasks, controllability, inclusion of Human--AI Collaborative (HAC) mechanisms, and whether the approach generates content from scratch. Controllability is defined as the approach's ability to adapt to user intent, provide fine-grained control over content generation, and support dynamic interaction during the writing process. HAC mechanisms involve iterative processes where human modifications to generated content are fed back into the model's \textit{learning} or \textit{generation}, enabling ongoing collaboration beyond one-off outputs without integrated feedback.}
        \vspace{-3mm}
        \label{tab:comparison}
        \begin{center}
            \resizebox{\textwidth}{!}{
            \begin{tabular}{@{}p{4.08cm} p{0.34cm} p{0.39cm} p{0.32cm} @{\hspace{1.8em}}p{5.7cm}@{\hspace{0.6em}}p{6.4cm}@{}}
                \toprule
                AI Generation Approach & ICL & Ctrl. & HAC & Quality Issues & Task \\
                \midrule
                PaperRobot~\citep{wang-etal-2019-paperrobot} & \ding{52} & \ding{52} & \ding{52} & Not LLM based, bad QA quality & Draft generation \\

                \citet{august-etal-2022-definition} & \ding{55} & \ding{55} & \ding{55} & Only definition & Scientific definition generation \\

                STORM~\citep{shao2024assisting}  & \ding{52} & \ding{55} & \ding{55} & Biased \& Factual hallucination & Article writing from scratch \\

                CO-STORM \citep{jiang2024unknown} & \ding{52} & \ding{55} & \ding{52} & Lack of consistency & Information-seeking assistance system \\

                CycleResearcher \citep{weng2024cycleresearcher}   & \ding{55} & \ding{55} & \ding{55} & Reward hacking \& Outdated & Paper generation and rating cycle \\

                \citet{ifargan2024autonomousllmdrivenresearchdata} & \ding{55} & \ding{52} & \ding{52} & From scratch & Automatic stepwise research \\

                Agent Lab~\citep{schmidgall2025agentlaboratory} & \ding{55} & \ding{55} & \ding{55} & Structure rigidity & Report, experimentation, and writing \\

                AI Scientist \citep{lu2024aiscientist} & \ding{55} & \ding{55} & \ding{55} & No control idea & Review and idea \\

                \citet{wang2025llmtimesmapreducev2entropydrivenconvolutionaltesttime} & \ding{55} & \ding{52} & \ding{52} & From scratch & Survey \\
                \bottomrule
                \texttt{XtraGPT} suite & \ding{52} & \ding{52} & \ding{52} & \textit{Controllable revision} & Paper revision \\
                \bottomrule
            \end{tabular}
            }
        \end{center}
        \vspace{-5mm}
    \end{table*}
\end{small}

We curate \texttt{ReviseQA} (Section~\ref{sec:instantiation}), a comprehensive dataset comprising 7,000 research papers from top-tier venues, annotated with 140,000 high-quality instruction--revision pairs. Building on \texttt{ReviseQA}, we introduce \texttt{XtraGPT}, the first family of \textbf{open-source} LLMs designed to support human-in-the-loop scientific writing, ranging from 1.5B to 14B parameters. Inspired by the modularity and interactivity of modern code editors~\citep{Cursor}, \texttt{XtraGPT} allows users to revise specific sections of a paper through instructions tied to writing goals (e.g. improving clarity, tightening motivation). By training on the criteria-guided \texttt{ReviseQA} dataset, \texttt{XtraGPT} internalizes the structural expectations and rhetorical strategies characteristic of top-tier publications. This enables the model to produce revisions that uphold academic rigor in alignment with real-world authoring workflows.

We validate the effectiveness of \texttt{XtraGPT} with both quantitative and qualitative evaluations. LLM-as-a-Judge results using length-controlled win rates show our revisions are consistently preferred over the original drafts. Model comparisons reveal that \texttt{XtraGPT} outperforms baselines, with the 7B variant matching GPT-4o-mini and the 14B variant surpassing it. Human evaluations confirm that \texttt{XtraGPT} produces rationale-aligned improvements that users are willing to adopt; furthermore, the significant rise in rating predictions implies that incorporating these revisions leads to measurable gains in overall quality, with the predicted overall rating increased 0.65$\pm$0.15 from 6.08 to 6.73 (p<0.001).

\section{Methodology}
\label{sec:framework}

\begin{figure*}[ht]
    \centering
    \includegraphics[width=0.98\linewidth]{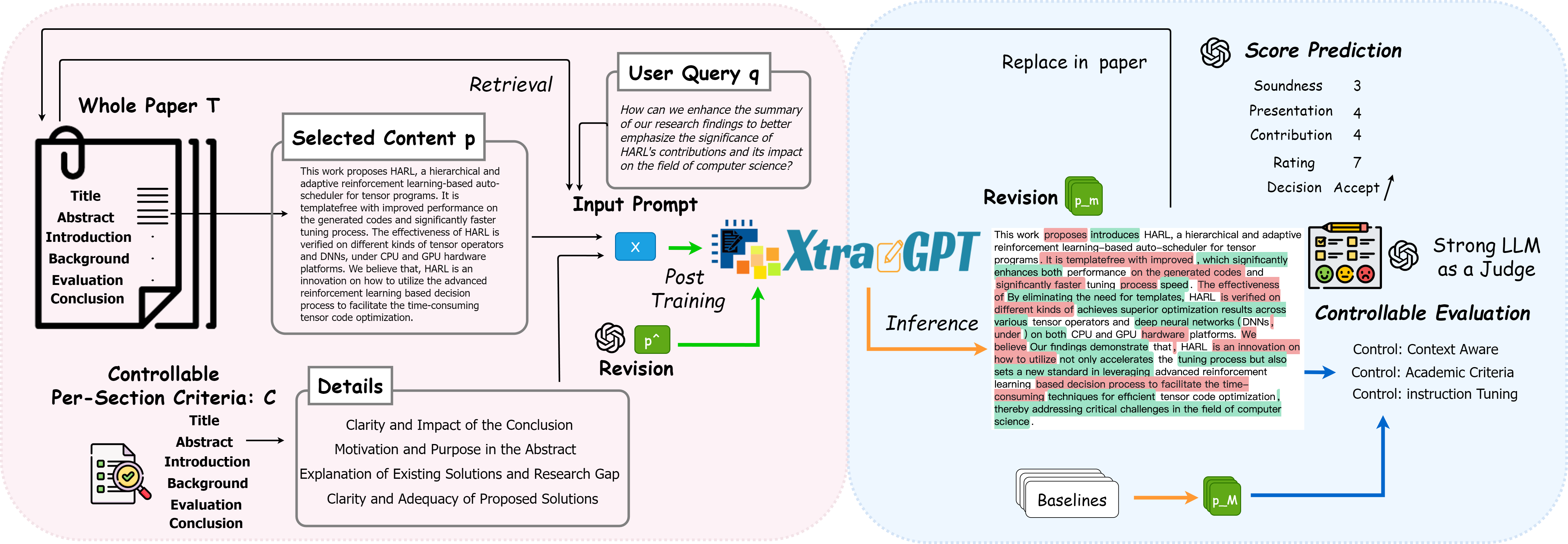}
    \caption{Overview. The post-training pipelines enable controllable, section-level, fine-grained paper revision.}
    \label{fig:xtra_main}
\end{figure*}

To enable LLMs to function as effective collaborators in academic writing, we propose a post-training framework designed to address the core challenges of context-awareness and controllability (in Figure \ref{fig:xtra_main}). This section outlines the problem formulation, the guiding principles of our framework, and the corresponding training objective.

\subsection{Problem Formulation}
The task of controllable academic paper revision can be formally defined as follows. Given a full paper draft $T$, a specific paragraph $p \subset T$ to be revised, and a natural language instruction $q$ from the author detailing the desired improvement, the goal is to generate a revised paragraph $\hat{p}$ that not only satisfies the instruction $q$ but also maintains coherence with the global context $T$. The model, parameterized by $\theta$, must learn the mapping:
\vspace{0mm}
\[
\hat{p} = \text{Model}_{\theta}(p, q, T)\vspace{-2mm}
\]

The key challenge lies in training $\text{Model}_{\theta}$ to use $(T, p, q)$, where $p$ is the human-selected span under the HAC protocol (Section~\ref{subsec:guiding-principles}), to produce context-consistent, targeted revisions.

\subsection{Guiding Principles}
\label{subsec:guiding-principles}
Our framework is built upon \textbf{Human-AI Collaboration (HAC)} as a foundational design philosophy \textit{rather than a standalone technical contribution}. HAC represents our conceptual stance (more in Section \ref{sec:ethics}): instead of pursuing full automation, which risks stifling originality, we cast the LLM as an assistant that augments authorial capabilities. In this paradigm, the human author directs the intellectual core (ideas, arguments, and initial drafts) while AI executes targeted, context-aware refinements based on specific instructions. This preserves authorial control and turns revision into an efficient, structured content.

\paragraph{HAC Protocol.} We concretize HAC as a three-step interface: (i) the human author \emph{selects} a target paragraph $p \subset T$ and \emph{issues} a natural-language instruction $q$; (ii) the model returns a revision $\hat{p}$ conditioned on $(T, p, q)$; (iii) the human \emph{reviews and integrates} $\hat{p}$, remaining the loop-closer across successive requests. Under this protocol, providing $p$ is \emph{not a task simplification but the interface itself}: authors typically know which passage to tighten and issue directed requests (e.g., ``clarify the motivation here''), rather than asking the model to autonomously locate weaknesses. The difficulty therefore lies not in localization but in producing a revision that is globally consistent with $T$ and faithful to $q$ at a fine granularity.

The philosophy is realized through two fundamental technical principles designed to make AI a more effective collaborator:

\paragraph{Criteria-Guided Intent Alignment.} Author instructions are often high-level, possibly muddled, and tied to established norms of academic writing (e.g., ``strengthen the contribution,'' ``clarify the methodology''). To make such instructions actionable for an LLM, our second principle is that \textbf{training data must be structured around a predefined set of academic writing criteria $\mathcal{C}$}. These criteria, derived from authoritative writing guides and reviewer guidelines, serve as a bridge between abstract authorial intent and concrete textual modifications. By training on instruction-revision pairs that are explicitly linked to these criteria (e.g., $q$ is an instantiation of a criterion $c \in \mathcal{C}$), the model learns to associate specific types of requests with corresponding revision strategies. This ensures \textit{controllability} by aligning the model's behavior with the structured, goal-oriented nature of writing.

\paragraph{Context-Aware Modeling.} To perform meaningful revisions, a model must understand a paragraph's function within the entire document. A requested change to the `Motivation` section in the introduction requires different considerations than a change to an `Analysis` paragraph in the evaluation section. Our framework's principle is that \textbf{the full document context $T$ must be an explicit input to the model during training and inference}. This forces the model to learn representations that are conditioned on the global narrative, structure, and terminology of the paper, enabling it to generate revisions that are contextually consistent and coherent. This principle directly facilitates \textit{in-context learning} by providing the necessary scope for the model to ground its revisions.

\begin{table}[ht]
    \centering
    \caption{Section-level revision criteria covering six key components of scientific papers.}
    \vspace{2mm}
    \label{tab:commentsdetails}
    \footnotesize
    \begin{tabular}{lp{0.48\linewidth}}
        \toprule
        \textbf{Aspect} & \textbf{Comments} \\
        \midrule
        \multirow{2}{*}{Title}
        & Consistency and Alignment of Title with Content \\
        & Conciseness and Clarity of Title \\
        \hline
        \multirow{4}{*}{Abstract}
        & Clarity and Impact of the Conclusion \\
        & Motivation and Purpose in the Abstract \\
        & Explanation of Existing Solutions and Research Gap \\
        & Clarity and Adequacy of Proposed Solutions \\
        \hline
        \multirow{6}{*}{Introduction}
        & Strength and Clarity of Motivation in the Introduction \\
        & Review of Existing Approaches in Introduction \\
        & Audience Alignment and Appropriateness \\
        & Clarity and Visibility of Contributions \\
        & Clarity and Specificity of Problem Definition \\
        & Integration of State-of-the-Art in Problem Framing \\
        \hline
        \multirow{3}{*}{Background}
        & Contextual Relevance and Clarity of Background \\
        & Coverage of Key Preliminary Concepts \\
        & Clarity and Consistency of Terminology \\
        \hline
        \multirow{3}{*}{Evaluation}
        & Experimental Setup Clarity and Reproducibility \\
        & Depth and Clarity of Figures and Tables Analysis \\
        & Experimental Support for Main Innovations \\
        \hline
        \multirow{2}{*}{Conclusion}
        & Broader Impact and Future Directions \\
        & Clarity and Impact of Key Innovations and Findings \\
        \bottomrule
    \end{tabular}
\end{table}
\subsection{Criteria-Guided Revision Data Curation}
Training effective models for scientific revision requires more than surface-level language correction; it necessitates a deep understanding of what constitutes a high-quality research paper. To simulate real-world revision needs and guide models toward meaningful improvements, we define a set of 20 section-level revision criteria, denoted by $\mathcal{C}$ (see Table~\ref{tab:commentsdetails}, Appendix \ref{sec:criteria_details}), spanning the six core components: Title, Abstract, Introduction, Background, Evaluation, and Conclusion. These criteria are grounded in authoritative scientific writing guidelines~\citep{widom2006tips} and have been further revised through expert revisions by experienced AI researchers.

Using these curated criteria, the instruction--revision pairs could target specific improvements in selected paragraphs, enabling LLMs to learn revision strategies aligned with established standards of scientific communication.

\subsection{Controllable Post-Training}
To operationalize these principles, we formulate a Controllable Post-Training (CPT) objective. We create a dataset $\mathcal{D}_{CPT}$ where each instance is a tuple $(q, T, p, \hat{p})$, curated according to the principles above. The training objective is to maximize the conditional log-likelihood of the target revision $\hat{p}$ given the original paragraph $p$, the instruction $q$, and the full paper context $T$:
\[
\mathcal{L}_{\text{CPT}}(\theta) = -\mathbb{E}_{(q, T, p, \hat{p})\sim \mathcal{D}_{CPT}} \left[ \log P_{\theta}(\hat{p} \mid q, T, p) \right]
\]
From the perspective of generative model alignment via high-quality demonstrations, we can observe the goal of optimization $P_{\theta}$ as
$
P_{\theta}(\hat{p} \!\mid\! q, T, p) \!=\! \sum_c P_{\theta}(\hat{p} \!\mid\! c, q, T, p) P_{\theta}(c \!\mid\! q, T, p)
$. Since the instruction-revision pairs $(q, \hat{p})$ in our dataset are generated based on specific criteria $c \in \mathcal{C}$, this objective implicitly trains the model to approximate the distribution $P(\hat{p} \mid q, T, p, c)$. This alignment with structured criteria is key to improving the model's controllability, its ability to follow instructions $q$, comply with section-level criteria $\mathcal{C}$, and maintain contextual consistency with $T$.

\subsection{Evaluation Protocol}

Previous work~\cite{peerreviewmt2024} has used \textit{surface-level} metrics such as ROUGE to evaluate full-text generation alignment in AI-assisted research. However, unlike tasks with a single ground truth, academic revision is inherently subjective. Such metrics only assess lexical overlap and fail to \textit{capture instruction intent adherence, controllable revision behavior, or overall, the semantic quality of improvements}. Given \textbf{the lack of established benchmarks} for controllable academic revision, we adopt preference-based evaluation, specifically the \textbf{Length-Controlled Win Rate (LC-Win Rate)}~\citep{dubois2024lengthcontrolled}, which allows for comparative evaluation while accounting for response length bias. Judgments are provided by an automatic evaluator, in our case, \texttt{alpaca\_eval\_gpt4\_turbo\_fn} \citep{zheng2023judgingllmasajudge}, which achieves 68.1\% agreement with human evaluations according to \citet{tatsu-lab-alpaca_eval}. We adapt the scoring prompts (Figure \ref{prompt4scoring}) and comparison prompts (Figure~\ref{prompt4judging}) to focus on controllability and alignment with instruction intent.

\paragraph{Single-Round as the HAC Unit.} Because the author is the loop-closer under our HAC protocol (Section~\ref{subsec:guiding-principles}), each evaluation instance measures one interaction step of the HAC workflow rather than a closed-loop autonomous rewrite. Cascading model outputs across rounds without the human in between would approximate full automation and contradict our design premise; cumulative, multi-step benefit instead appears in the paper-level study (Section~\ref{sec:real_world_hac}, Figure~\ref{fig:combined-results}).

\paragraph{Addressing Length Bias in LLM Judges.}
While win rate is a commonly used metric for evaluating model performance in paragraph rewriting tasks, it becomes unreliable in the presence of length bias, a phenomenon well-documented in prior work, where LLM judges tend to prefer longer responses over shorter ones~\citep{koo2024benchmarking,shen2023loose}. Given the substantial variation in output lengths across models (as shown in Table~\ref{tab:model_output_length}), the \textbf{length-controlled win rate} \citep{dubois2024lengthcontrolled} should be ensured to ensure fairer comparisons. This metric explicitly adjusts for length differences, mitigating the bias introduced by verbosity. Our approach is further supported by recent evaluation protocols such as AlpacaEval \citep{tatsu-lab-alpaca_eval}, and enables more accurate assessment of the true quality of model-generated revisions.

\paragraph{Why use LLM-as-a-Judge for controllable paper revision?}
For subjective sequence-level tasks such as paragraph rewriting, LLM-based evaluation provides consistent and scalable feedback (e.g., on controllability). This practice has proven effective in the development of models such as InstructGPT and ChatGPT. Prior work has shown that LLM judges align well with human reviewers in research settings~\citep{lu2024aiscientist}. Compared to manual annotation, this approach offers faster feedback while preserving evaluation quality.

\paragraph{Length-Controlled Win Rate Formulation.}
Let $m$ denote the target model and $M$ denote a baseline model. Let $\theta$ be the raw prediction score, and let $z_m$ and $z_M$ denote their respective outputs given input $x$. The length-controlled win rate is computed as:
\[
q_{\theta, \phi, \psi}(y = m \mid z_m, z_M, x) := \text{logistic}
\left( \underbrace{\theta_m - \theta_M}_{\text{model term}} + \underbrace{\phi_{M,b} \cdot \tanh \left( \frac{\text{len}(z_m) - \text{len}(z_M)}{\text{std}(\text{len}(z_m) - \text{len}(z_M))} \right)}_{\text{length term}} \right)
\]
The instruction difficulty term in this work is omitted to focus solely on the effectiveness of the revisions. The final LC win rate is computed as:
\vspace{-1mm}
\[
\text{winrate}^{LC}(m, M) = 100 \cdot \mathbb{E}_x \left[ q_{\theta, \phi, \psi}(y = m \mid z_m, z_M, x) \right]
\vspace{-1.5mm}
\]

This formulation adjusts for output length variation, ensuring that model comparisons reflect actual quality improvements rather than superficial verbosity.

\section{Instantiation}
\label{sec:instantiation}

To validate the framework proposed in Section \ref{sec:framework}, we conducted a large-scale study. This section details the specific implementation of our data curation process for model training, designed to serve as a concrete testbed for our methodology. Beyond standalone release, \texttt{XtraGPT} powers \textit{PaperDebugger}~\cite{hou2026paperdebugger} as the refiner, a plugin providing real-time refinement directly within Overleaf.

\paragraph{ReviseQA}
Following the principles outlined in our framework, we created \texttt{ReviseQA}, the first large-scale dataset designed for controllable, context-aware paper revision. The curation process itself is a simulation of our envisioned \textbf{Human-AI Collaboration workflow}, designed to capture the essence of iterative, instruction-driven refinement.

\paragraph{Source Data Acquisition and Context Modeling.}
To ensure the data reflects high-quality, contemporary scientific writing, we begin by collecting all research papers submitted to ICLR 2024 via the OpenReview API. Faced with practical constraints on time and computation, this selection was motivated by ICLR's high paper quality and data transparency, and the timeliness of the 2024 dataset, as its guidelines had been updated and 2025 data was not yet released during our project. Review statistics, including soundness, presentation, contribution, and overall ratings, are shown in Figure~\ref{fig:iclr24distributions}, and the distribution of paper lengths is presented in Figure~\ref{fig:tokendistribution}. Among the submissions, 64.71\% received reviewer replies, with 82.4\% of those discussions leading to a final decision. Based on the parsed and filtered PDF files, the overall acceptance rate was 36.3\%.

After excluding 40 out of 7,040 excessively long papers, we retained 7,000 valid submissions. Each PDF was converted into structured markdown using \textsc{Nougat}~\citep{blecher2023nougat_pdfparser}, a deep learning-based academic parser that outperforms prior rule-based approaches, as demonstrated by \citet{li2024readdoc_paparparserqualityassurance}. The resulting set of tokenizable paper texts is denoted as $\mathcal{T}$.

To ensure length consistency and content relevance, we applied a post-processing step to remove non-essential sections, such as acknowledgments and references, retaining only the main body of the paper. This allowed $\mathcal{T}$ to remain within the maximum token limit of 16,384 tokens, making it suitable for downstream modeling. To adhere to our principle of \textit{Context-Aware Modeling}, we retained the main body of each paper as the full context $T$, providing the AI with the same global view a human collaborator would have.

\paragraph{Simulating Collaborative Interaction via Criteria-Guided Curation.}
Using these curated criteria in Table \ref{tab:commentsdetails}, we generate instruction--revision pairs that target specific improvements in selected paragraphs, enabling LLMs to learn revision strategies aligned with established standards of scientific communication. For each article $T\in \mathcal{T}$, we sample user-identifiable paragraphs from the aforementioned six core sections and generate 20 criterion-guided questions using carefully constructed prompts. For each sampled paragraph $p\in T$, we first generate an instruction query $q$ based on a selected criterion $c\in\mathcal{C}$. Given the full paper context $T$, we then use GPT-4o-mini, a model with a reported hallucination rate of only 1.7\% \citep{hallucationleaderboard}, to generate the corresponding revised paragraph, denoted as $\hat{p}$. We selected GPT-4o-mini for data generation instead of more recent state-of-the-art LLMs (e.g. DeepSeek R1~\citep{deepseekai2025deepseekr1}) because our task does not require complex reasoning, multi-step planning, or deep domain knowledge. Rather, it emphasizes long-context understanding and localized revision, areas in which GPT-4o-mini performs reliably. As validated by human annotators (see Table~\ref{tab:baseline_and_human_alignment}), GPT-4o-mini produces coherent and high-quality paragraph-level revisions. The full set of criteria is illustrated in Figures~\ref{fig:criteria_title} through~\ref{fig:criteria_conclusion}, and the prompt design is detailed in Figure~\ref{prompt4GENQA}.

\begin{table}[t]
    \centering
    \caption{Human evaluation of improvement acceptance rates before and after revision. Three human evaluators assessed 5, 3, and 5 papers respectively, scoring 100, 60, and 100 instruction--revision pairs using a 1--5 scale. The \textbf{Aggregated} column reports the average scores across all three evaluators.}
    \vspace{2mm}
    \label{tab:baseline_and_human_alignment}
    \footnotesize
    \setlength{\tabcolsep}{3pt}
    \begin{tabular}{l ccc|c}
        \toprule
        \textbf{Criterion} & \textbf{Judge 1} & \textbf{Judge 2} & \textbf{Judge 3} & \textbf{Aggregated} \\
        \midrule
        \multicolumn{5}{l}{\textbf{GPT-4o-Mini}} \\
        -Instruction Following & 3.83 & 3.72 & 3.87 & 3.81 \\
        -Criteria Alignment & 3.68 & 3.74 & 3.83 & 3.75 \\
        -In-Context Reference & 2.97 & 3.32 & 3.64 & 3.31 \\
        -Revision Acceptance & 2.46 & 3.09 & 3.58 & 3.04 \\
        \midrule
        \multicolumn{5}{l}{\textbf{GPT-o1-mini}} \\
        -Instruction Following & 3.82 & 3.95 & 3.74 & 3.84 \\
        -Criteria Alignment & 3.70 & 3.85 & 3.71 & 3.76 \\
        -In-Context Reference & 3.10 & 3.40 & 3.68 & 3.40 \\
        -Revision Acceptance & 2.80 & 3.32 & 3.64 & 3.25 \\
        \bottomrule
    \end{tabular}
\end{table}

These pairs form the foundation of our \texttt{ReviseQA} dataset, which enable models to learn principled, context-aware revisions grounded in realistic scientific writing goals. In total, this process yields 140,000 high-quality instruction--response pairs. To facilitate rigorous evaluation, we construct a held-out QA benchmark by randomly sampling 5\% of the papers and their corresponding instruction-revision pairs, totaling 350 papers and 7,000 QA pairs. Our benchmark supports model comparison using length-controlled win rate evaluation~\citep{dubois2024lengthcontrolled}, with \texttt{XtraGPT} serving as the anchor model.

We formally define the dataset used for Contollable Post-Training (CPT) as $\mathcal{D}_{CPT}$, where each data instance $D\in\mathcal{D}_{CPT}$ is in the form of $D\in\mathcal{D}_{CPT} = \{(q, T, p, \hat{p})\}$.

\paragraph{Data Quality Validation.}
\label{sec:quality_validation}

Throughout the data collection and generation process, we applied strict quality control procedures to ensure the reliability and utility of the dataset for both training and evaluation. In particular, we conducted a human evaluation to verify that the revisions are applicable and effective across papers with diverse writing styles and technical focuses spanning multiple research domains.

Three Ph.D. students in computer science (referred to as Judges 1--3) were recruited as evaluators. Each annotator was assigned between 60 and 100 instruction--revision pairs randomly sampled from \texttt{ReviseQA}. Each revision is labeled according to the criteria defined in Figure~\ref{fig:criterias}, focusing on four key aspects: \textbf{(1) Instruction Following:} Whether the revision correctly follows the given instruction. \textbf{(2) Criteria Alignment:} Whether the revised text improves the original content based on the predefined section-level criteria (Table~\ref{tab:commentsdetails}). \textbf{(3) In-Context Reference:} Whether the output appropriately references relevant information within the selected paragraph or surrounding context. \textbf{(4) Revision Acceptance:} Whether the revision is compelling enough for the evaluator to prefer it over the original.

Evaluation results in Table \ref{tab:baseline_and_human_alignment} show that the revisions are consistently effective across domains, supporting the validity of \texttt{ReviseQA} as a high-quality resource for instruction-based writing revision.

\section{Experiment}
\label{sec:experiment}
We conduct experiments on \texttt{ReviseQA} to answer the following research questions:
\begin{itemize}[leftmargin=*]
    \item \textbf{Effectiveness} (Section \ref{sec:revise_comparison}): How does \texttt{XtraGPT} perform in comparison to representative LLM baselines?
    \item \textbf{Adaptability} (Section \ref{sec:revise_comparison}): Can \texttt{XtraGPT} be effectively integrated with different architectures and sizes?
    \item \textbf{Human Preference} (Section \ref{sec:qualitative_eval}): How are the revisions produced by \texttt{XtraGPT} perceived in terms of quality and usefulness by human evaluators?
    \item \textbf{Real-World Applicability} (Section \ref{sec:real_world_hac}): How well does \texttt{XtraGPT} support practical human--AI collaborative paper revision workflows?
\end{itemize}

\subsection{Experimental Setup}
\paragraph{Benchmarking.} We benchmark representative LLMs on \texttt{ReviseQA} to evaluate controllable paper revision capabilities. During preliminary analysis, we identified several common issues with LLM-based revisions: overuse of generic GPT-style language (e.g., ``comprehensive'') that artificially inflates the perceived impact of the paper, superficial edits that do not address the core revision intent, and a tendency to generate unnecessarily long revised segments. To mitigate these issues, we carefully designed our generation prompt (Figure~\ref{prompt4GENQA}) to encourage focused and contextually grounded improvements. Additionally, we introduced evaluation-specific prompts (Figures~\ref{prompt4judging} and~\ref{prompt4scoring}) that guide models and human judges to assess revisions with an emphasis on clarity, conciseness, and content relevance.

\paragraph{\texttt{XtraGPT} Training Setup.} \texttt{XtraGPT} models are designed of different sizes: 1.5B (based on \texttt{Qwen-2.5-1.5B-Instruct}), 3.8B (based on \texttt{phi3.5-3.8b}), 7B (based on \texttt{Qwen-2.5-7B\\-Instruct}), and 14B (based on \texttt{phi4-14b}). Full CPT training (outperforming LoRA~\cite{hu2022lora}, see Appendix~\ref{sec:lora_tuned}) was conducted using 4 NVIDIA H100 GPUs, each with 80 GB of memory. For inference on \texttt{ReviseQA}, we employed the vLLM framework~\citep{kwon2023vllm} on a system with a single NVIDIA A100 GPU (80 GB). The computing environment was configured with CUDA 12.2 and cuDNN 9.1 to ensure optimal deep learning performance. Hyperparameter details are provided in Appendix~\ref{sec:hypers}.

\begin{table*}[t]
\footnotesize
\centering
\caption{Length-controlled (LC) win rates of various models against the 7B version of \texttt{XtraGPT} (used as the \textit{anchor}) across different evaluation categories. Models above the grey anchor bar surpass \texttt{XtraGPT}; those below under-perform. Evaluations are conducted using a modified \texttt{alpaca\_eval\_gpt4\_turbo\_fn} judge with the prompt shown in Figure~\ref{prompt4judging}.}
\label{tab:lc_win_rate}
\resizebox{\textwidth}{!}{
\setlength{\tabcolsep}{3pt}
\begin{tabular}{lcccccccc}
\toprule
\textbf{Models} & \textbf{Title} & \textbf{Abstract} & \textbf{Introduction} & \textbf{Background} & \textbf{Evaluation} & \textbf{Conclusion} & \textbf{Overall}$\uparrow$\\
\midrule
% \textbf{Phi-4 (14B)} & 16.87 & 58.55 & 66.11 & 75.07 & 80.63 & 51.48 & 61.73 \\
% Qwen2-72B-Instruct & * & * & * & * & * & * & * \\
QwQ-32B & 46.58 & 85.34 & 81.99 & 83.82 & 82.64 & 95.69 & 80.86 \\
DeepSeek-v3-671B & 56.42 & 65.71 &  68.32 & 74.12 & 72.11 & 64.83 & 67.70 \\
Qwen2.5-72B-Instruct & 50.35 & 56.78 & 64.75 & 67.13 & 64.59 & 60.34 & 61.61 \\
XtraGPT-14B & 55.29 & 59.43 & 50.90 & 59.43 & 57.87 & 52.11 & 55.49 \\
GPT-4o-Mini & 48.80 & 47.43 & 55.73 & 66.07 & 45.67 & 39.03 & 51.75 \\
\rowcolor{lightgray}
\multicolumn{8}{c}{\texttt{XtraGPT}-7B; base: \texttt{Qwen-2.5-7B-Instruct}(\textit{anchor}$\uparrow$)} \\
% Qwen-2.5-7B-Instruct & 50.41 & 47.11 & 43.71 & 46.56 & 46.05 & 49.79 & 46.44 \\
% Qwen-2.5-7B-Instruct (10\%) & 50.48 & 38.44 & 42.06 & 53.33 & 46.82 & * & * \\
% Qwen2.5-7B-Instruct-paperfull214/checkpoint-1000-new & 51.85 & 50.32 & 44.85 & 44.44 & 46.84 & 50.00 & * \\
Qwen2.5-7B-Instruct & 39.93 & 45.14 & 45.64 & 39.28 & 33.87 & 31.17 & 40.80 \\
Phi-4 & 4.57 & 44.78 & 50.90 & 52.54 & 26.33 & 42.01 & 40.71 \\
% \textbf{DeepSeek-R1-Distill-Llama3-8B}\\
% \textbf{qwen-1.5-110B}\\
% Qwen-2.5-7B-Instruct & 46.18 & 63.43 & 45.74 & 46.90 & 46.44 & 49.50 & \textcolor{red}{49.98} & 49.70 \\
Qwen-QWQ-32B-Preview & 37.83 & 34.57 & 32.13 & 40.58 & 30.04 & 32.91 & 34.22 \\
Llama-3.1-8B-Instruct & 34.78 & 30.64 & 35.31 & 41.60 & 40.29 & 18.36 & 33.51 \\
DeepSeek-R1-Distill-Llama-8B & 37.98 & 15.19 & 10.87 & 9.64 & 9.73 & 9.38 & 13.94 \\
% XtraGPT-1.5B & 41.91 (*) & 43.54 (*) & & 39.19 (*) & 42.41 (*) & 35.53 (*) &\\
Qwen2.5-1.5B-Instruct & 36.07 & 30.87 & 25.80 & 21.34 & 24.18 & 24.27 & 26.80 \\
GPT-3.5-Turbo & 25.73 & 23.99 & 21.52 & 23.16 & 30.97 & 17.39 & 24.24 \\
% DeepSeek-R1-Distill-Qwen-7B (deepseek show bad ins-following!) & 23.26 & 50.15 & 51.45 & 52.83 & 64.78 & 54.60 & 50.89 \\
Llama-3.2-3B-Instruct & 19.93 & 6.45 & 9.35 & 3.80 & 8.26 & 4.64 & 8.73 \\
% DeepSeek-R1-Distill-Qwen-7B & 4.30 & 5.03 & 0.75 & 0.84 & 0.00 & 3.63 & 2.15 \\
\bottomrule
\end{tabular}}
\end{table*}

\begin{table*}[t]
\footnotesize
\centering
\caption{Length-controlled (LC) win rates of various models against \texttt{XtraGPT} (\textit{anchor}) across evaluation categories. Models are ranked in descending order based on their weighted LC win rates. A modified version of \texttt{alpaca\_eval\_gpt4\_turbo\_fn} was employed as judge  (see Prompt~\ref{prompt4judging}).}
\label{tab:adaptability}

\begin{footnotesize}
\resizebox{\textwidth}{!}{
\setlength{\tabcolsep}{3pt}
\begin{tabular}{lcccccccc}
\toprule
\textbf{Models} & \textbf{Title} & \textbf{Abstract} & \textbf{Introduction} & \textbf{Background} & \textbf{Evaluation} & \textbf{Conclusion} & \textbf{Overall}$\uparrow$\\
\midrule
\rowcolor{lightgray}
\multicolumn{8}{c}{\texttt{XtraGPT}-7B (\textit{anchor}$\uparrow$)} \\
Qwen2.5-7B-Instruct & 39.93 & 45.14 & 45.64 & 39.28 & 33.87 & 31.17 & 40.80 \\
\rowcolor{lightgray}
\multicolumn{8}{c}{\texttt{XtraGPT}-14B(\textit{anchor}$\uparrow$)} \\
Phi-4 (14B) & 5.09 & 40.11 & 39.63 & 50.00 & 24.27 & 39.08 & 35.47 \\
\rowcolor{lightgray}
\multicolumn{8}{c}{\texttt{XtraGPT}-3.8B (\textit{anchor}$\uparrow$)} \\
Phi-3.5-mini-instruct (3.8B) & 17.52 & 43.74 & 31.64 & 34.41 & 25.85 & 58.24 & 34.86 \\
\rowcolor{lightgray}
\multicolumn{8}{c}{\texttt{XtraGPT}-1.5B(\textit{anchor}$\uparrow$)} \\
Qwen2.5-1.5B-Instruct & 16.83 & 10.70 & 10.34 & 5.41 & 4.39 & 15.81 & 9.98 \\
\rowcolor{lightgray}
\multicolumn{8}{c}{\texttt{XtraGPT}-3B (\textit{anchor}$\uparrow$)} \\
Llama-3.2-3B-Instruct & 0.14 & 0.57 & 0.56 & 0.38 & 0.26 & 1.89 & 0.58 \\
\bottomrule
\end{tabular}}
\end{footnotesize}

\end{table*}
\subsection{Comparison with Paper Revision Models} \label{sec:revise_comparison}

\begin{table}[h]
\centering
\fontsize{7}{8}\selectfont
\begin{tabular}{p{0.34\linewidth} p{0.61\linewidth}@{}}
\toprule
\multicolumn{2}{p{0.98\linewidth}}{\textbf{Shared Content (from abstract):} ... To achieve better generalizability of learned graph representation, we innovatively propose a \textbf{\underline{C}}onstrained \textbf{\underline{V}}ariational \textbf{\underline{G}}eneration (CVG) framework to generate generalizable graphs. Our framework is implemented based on the Variation Graph Auto-Encoder (VGAE) structure and optimized under the guidance of the Graph Information Bottleneck (GIB) principle, with its effectiveness validated by our theoretical analysis. We conduct extensive experiments on real-world datasets and demonstrate the superiority of our framework over state-of-the-art baselines.} \\
\multicolumn{2}{p{0.98\linewidth}}{\textbf{Question:} How can we effectively summarize our key innovations and their contributions to the field of Generalizable Graph Learning? What specific results and comparisons to existing methods should we include to demonstrate the significance of our findings? how the experiments reflect the effectiveness of our framework, for example improve how many percent on which task?} \\
\midrule
\textbf{GPT-4o Response:} & \textbf{XtraGPT Response:} \\
\midrule
... To tackle this, we propose a method that not only identifies candidate reward functions but also ensures that the selected reward functions are aligned with the underlying task. We establish the conditions under which PAGAR can guarantee successful task completion, thereby providing a robust framework for enhancing the reliability of IRL-based IL systems. Our experimental results demonstrate that PAGAR significantly outperforms existing baselines across various complex and zero-shot IL tasks, showcasing its effectiveness in mitigating reward misalignment.  & ... To achieve better generalizability of learned graph representation, we innovatively propose a \textbf{\underline{C}}onstrained \textbf{\underline{V}}ariational \textbf{\underline{G}}eneration (CVG) framework to generate generalizable graphs. Our framework is implemented based on the Variation Graph Auto-Encoder (VGAE) structure and optimized under the guidance of the Graph Information Bottleneck (GIB) principle. The key innovations of our CVG framework include:
Constrained Variational Generation: By leveraging the VGAE structure, our framework generates graphs that are constrained to maintain essential invariant properties while allowing flexibility to adapt to unseen domains.
Graph Information Bottleneck (GIB) Optimization: The GIB principle guides the optimization process to ensure that the generated graphs retain the most informative features for classification while discarding irrelevant variations.
Our theoretical analysis validates the effectiveness of these innovations, ensuring that the learned representations are robust to distribution shifts.\\
\midrule
\textbf{Weaknesses:} \textit{No length control}; \textit{Simply paraphrase}; \textit{Generalized description}; \textit{Single interaction, lack of context}; \textit{Like chatbot, not HAC.} & \textbf{Strengths:} \textit{Holistic Processing, e.g.\ refers to experiment results}; \textit{Critical feedback}; \textit{Without spoiling the flow of thought}; \textit{Replace back to paper automatically.} \\
\bottomrule
\end{tabular}
\vspace{2mm}
\caption{Case study comparing GPT-4o and XtraGPT.}
\label{tab:casestudy}
\end{table}

As shown in the length-controlled win rate results in Table~\ref{tab:lc_win_rate}, \textbf{\texttt{XtraGPT}-7B revisions are consistently preferred over same-scale (7B/8B) open-source models}. Moreover, \texttt{XtraGPT}-7B outperforms GPT-4o-mini\footnote{GPT-4o-mini is for benchmark construction and evaluation; GPT-4o serves as the SOTA reference for case studies.} on several sections, including the abstract, evaluation, and conclusion. When scaling the model to 14B parameters, the larger \texttt{XtraGPT} variant ranks higher on the leaderboard than GPT-4o-mini, demonstrating strong performance at scale. A case study in Table~\ref{tab:casestudy} further highlights \texttt{XtraGPT}-7B's practical applicability over GPT-4o's by showcasing nuanced improvements beyond numerical metrics.

To evaluate the generalizability of our framework, and as an implicit ablation study on data curation, we trained and compared models based on representative open-source LLM backbones of varying sizes, ranging from 1.5B to 14B parameters. The corresponding results are reported in Table~\ref{tab:adaptability}. Across all configurations, \texttt{XtraGPT} consistently outperforms its respective base model. Base models rely on general-purpose capabilities from broad pre-training, whereas \texttt{XtraGPT} is fine-tuned on our specialized dataset, structured by academic writing criteria. The significant improvement demonstrates that \textbf{criteria-guided fine-tuning yields substantial gains over general abilities, validating the effectiveness of our approach.}

\subsection{Qualitative Evaluation}
\paragraph{Human Perception of Revision Quality.} \label{sec:qualitative_eval}
\begin{table}[h]
\footnotesize
\centering
\setlength{\tabcolsep}{3pt}
\caption{Effectiveness of \texttt{XtraGPT} as validated by human ratings on a 1--5 scale.}
\label{tab:xtragpt_and_human_alignment}
\begin{tabular}{p{3cm}ccc|c}
\toprule
\texttt{XtraGPT} & \textbf{Judge 1} & \textbf{Judge 2} & \textbf{Judge 3} & \textbf{Aggregated} \\ \midrule
Instruction Following & 3.25 & 3.99 & 4.09 & 3.78 \\
Criteria Alignment & 3.34 & 3.70 & 4.09 & 3.71 \\
In-Context Reference & 2.80 & 3.40 & 4.06 & 3.42 \\
Revision Acceptance & 2.46 & 3.23 & 4.01 & 3.23 \\ \hline
\end{tabular}
\end{table}

Beyond the length-controlled win rate results presented in Table~\ref{tab:lc_win_rate} and Table~\ref{tab:adaptability}, we further assess the effectiveness of revisions through human evaluation by measuring users' willingness to adopt the generated revision in place of the original contents.

Following a similar evaluation setup to our dataset quality validation (Section~\ref{sec:quality_validation}), we engaged three evaluators with research backgrounds and provided them with detailed criteria for assessing model performance across four key dimensions. These criteria are visualized in Figure~\ref{fig:criteria_title},\ref{fig:criteria_abstract},\ref{fig:criteria_introduction},\ref{fig:criteria_background},\ref{fig:criteria_evaluation},\ref{fig:criteria_conclusion}. Table~\ref{tab:xtragpt_and_human_alignment} reports the human-assigned quality ratings of \texttt{XtraGPT} revisions, demonstrating that the outputs are consistently perceived as high-quality and aligned with human expectations. Furthermore, Table \ref{tab:casestudy} shows a concrete comparison of \texttt{XtraGPT}-revised paper excerpt and the LLM-revised paper excerpt.

\subsection{Reliability of Human--AI Collaborative Revision}
\label{sec:real_world_hac}

To assess the effectiveness of our paper improvement tool, we employed \textsc{AI-Scientist} \cite{lu2024aiscientist}, an open-source system developed by Sakana AI. \textsc{AI-Scientist} includes a functionality that simulates the peer review process and provides \textbf{paper-level} evaluations. Notably, the system's capability is underpinned by the success in generating a fully AI-authored paper that passed peer review at a top-tier machine learning workshop \cite{yamada2025aiscientistv2}. Before applying the system to our own data, we first establish the reliability of \textsc{AI-Scientist} as an evaluator of scientific writing quality.

\begin{table}[h]
\footnotesize
    \centering
    \caption{Performance of \texttt{XtraGPT} on predicting decisions.}
    \vspace{1mm}
    \label{tab:ai-scientist-results}
    \begin{tabular}{lc}
    \toprule
    \textbf{Metric} & \textbf{Value (\%)} \\
    \midrule
    Accuracy & 70.57 \\
    Precision (Accept) & 57.14 \\
    Recall (Accept) & 80.00 \\
    F$_1$ Score (Accept) & 66.64 \\
    \bottomrule
    \end{tabular}
\end{table}
To this end, we validated \textsc{AI-Scientist} by applying it to a sample of 54 ICLR 2024 submissions with known ground-truth acceptance decisions. The system was tasked with predicting the acceptance outcome for each paper. Results indicate that \textsc{AI-Scientist} achieves a reasonable level of accuracy and reliability in approximating peer review judgments, thereby justifying its use as an automated evaluator in our study. A summary of these results is provided in Table~\ref{tab:ai-scientist-results}.

\begin{figure*}[ht]
    \centering
    \begin{minipage}[b]{0.55\textwidth}
        \centering
        \includegraphics[width=\linewidth]{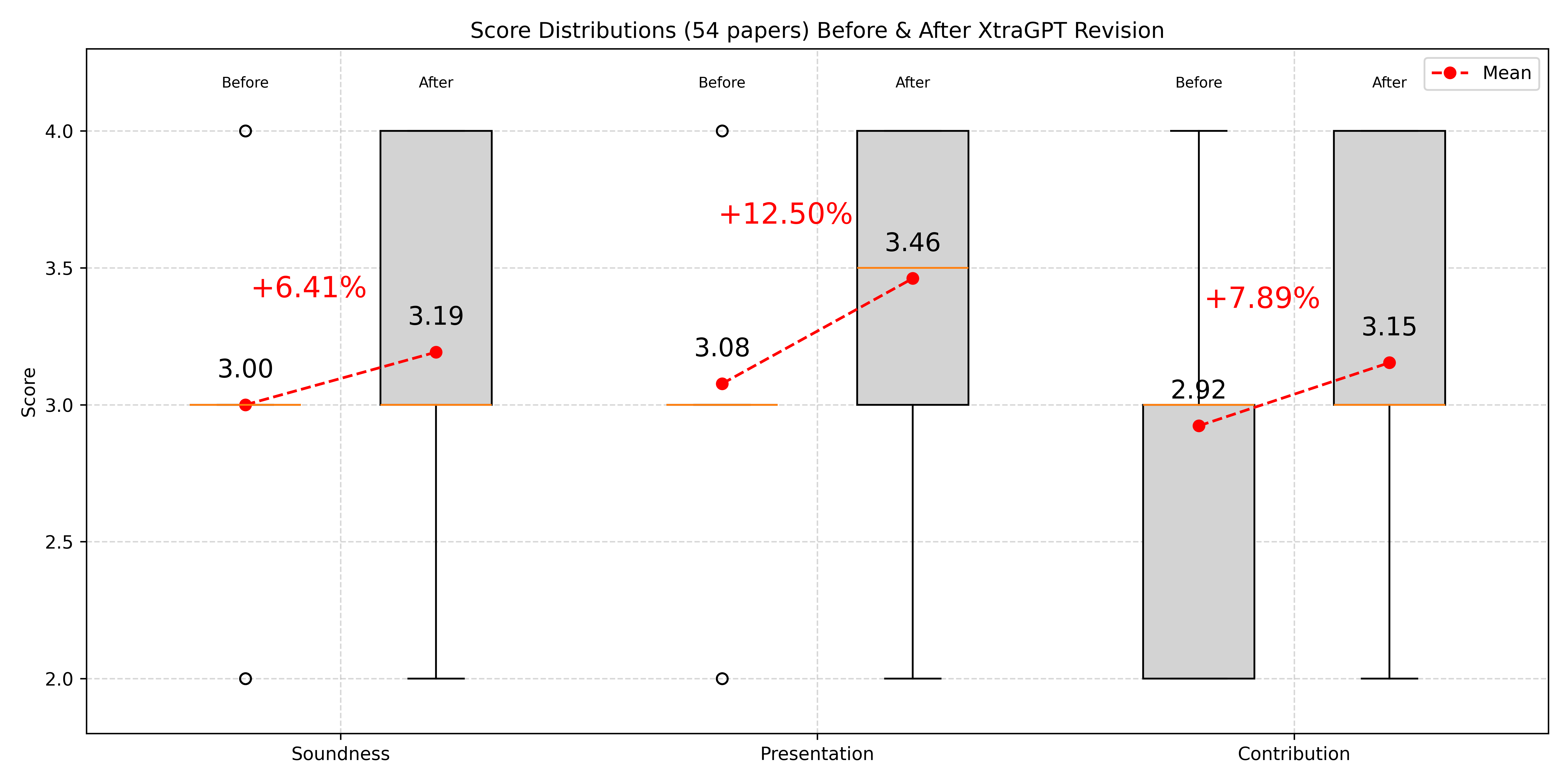}
    \end{minipage}
    \begin{minipage}[b]{0.4\textwidth}
        \centering
        \includegraphics[width=\linewidth]{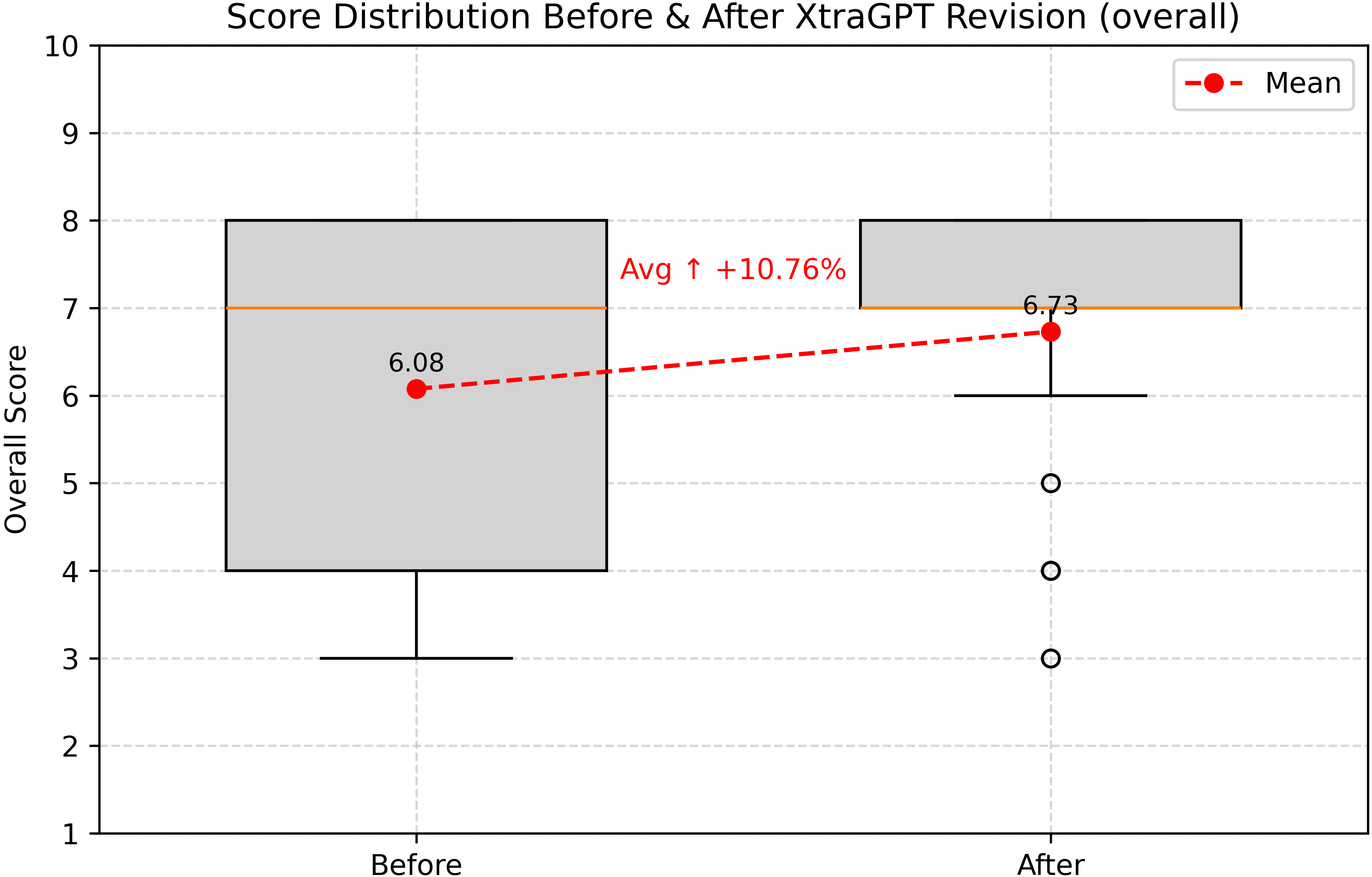}
        \vspace{-2.5mm}
    \end{minipage}
    \vspace{-1mm}
\caption{Paper quality scores and overall ratings from o1-based \textsc{AI-Scientist}, before and after \texttt{XtraGPT} revision. \textbf{Left:} Evaluation of revision quality. On average, contribution scores increased by 7.89\%, presentation by 12.50\%, and soundness by 6.41\%. \textbf{Right:} Distribution of overall ratings on a 1--10 scale before and after revision. Average rating increased by 10.76\%.}
    \label{fig:combined-results}
    \vspace{-3mm}
\end{figure*}

Following the validation of \textsc{AI-Scientist}'s reliability, we employed it to evaluate the full-paper quality improvements made by \texttt{XtraGPT}. Specifically, \textsc{AI-Scientist} was used to assign scores on three core dimensions from the review rubric: \textit{contribution}, \textit{presentation}, and \textit{soundness}, each rated on a 1--4 scale, and \textit{overall rating} on a 1--10 scale. For each of the 54 papers in our evaluation set, we obtained these scores both before and after revision by \texttt{XtraGPT} (repeat 3 times for t-test). As shown in the left of Figure~\ref{fig:combined-results}, \texttt{XtraGPT} yielded consistent gains across all three dimensions. For evaluation of revision quality, contribution (1--4 scale) raises 0.23$\pm$0.10 from 2.92 to 3.15 (p<0.001). Presentation (1--4 scale) raises 0.38$\pm$0.10 from 3.08 to 3.46 (p<0.001). Soundness (1--4 scale) raises 0.19$\pm$0.08 from 3.00 to 3.19 (p=0.004). In addition to these criteria-specific improvements, the overall score---which ranges from 1 to 10 and reflects the holistic recommendation---also improved, increased 0.65$\pm$0.15 from 6.08 to 6.73 (p<0.001) as shown in the right of Figure~\ref{fig:combined-results}. These suggest that \texttt{XtraGPT}'s revisions lead to measurable improvements in dimensions that are central to peer review. Crucially, as these revisions are grounded in \textit{original human-authored} drafts, they are consistently \textbf{classified as human-written} under both Fast-DetectGPT \cite{bao2023fast} and Binoculars \cite{hans2024binoculars} (Appendix ~\ref{app:ai_detection}), further validating the effectiveness of our collaborative framework.

\begin{table*}[t]
\centering
\caption{Ablation Study. Length-Controlled Win Rates against \texttt{XtraGPT}-7B (Anchor).}
\label{tab:ablation_study}
\begin{footnotesize}
\resizebox{\textwidth}{!}{
\setlength{\tabcolsep}{3pt}
\begin{tabular}{lcccccccc}
\toprule
\textbf{Models} & \textbf{Title} & \textbf{Abstract} & \textbf{Introduction} & \textbf{Background} & \textbf{Evaluation} & \textbf{Conclusion} & \textbf{Overall}$\uparrow$\\ 
\midrule
\rowcolor{lightgray}
\multicolumn{8}{c}{\texttt{XtraGPT}-7B (\textit{anchor}$\uparrow$)} \\
Post-Trained w/o Criteria & 50.41 & 45.11 & 41.71 & 44.56 & 44.05 & 47.79 & 44.65 \\
Qwen-2.5-7B-Instruct (Base) & 39.93 & 45.14 & 45.64 & 39.28 & 33.87 & 31.17 & 40.80 \\
Post-Trained w/o Context ($T$) & 37.14 & 33.83 & 36.43 & 40.74 & 40.08 & 11.76 & 34.71 \\
\bottomrule
\end{tabular}}
\end{footnotesize}
\end{table*}

\subsection{Ablation Study}
\label{sec:ablation}

To investigate the contribution of each component within controllable post-training (CPT), we conduct an ablation study (Table~\ref{tab:ablation_study}) to summarize the performance gaps relative to the \texttt{XtraGPT} anchor when specific components are removed. 

\noindent \textbf{Impact of Criteria $\mathcal{C}$:} 
Removing criteria (w/o Criteria) variant falls short of the full CPT (44.65 vs. 50.00). It suggests that generic post-training alone is insufficient to reach CPT, particularly in structured sections like the Introduction and Abstract, where criteria-guided alignment is important.

\noindent \textbf{Impact of Context $T$:} Omitting document context (\(T\)) results in the most substantial degradation relative to CPT (34.71), with the \textit{Conclusion} section suffering a near-total collapse (11.76). This suggests that context-awareness is integral to generating coherent, full-paper revisions.

\noindent \textbf{Length Control:} We treat length control as a necessary evaluation protocol. Consistent with \citet{dubois2024lengthcontrolled}, it is essential to mitigate verbosity bias and reveal true quality improvements, as evidenced by the discrepancy between raw and length-controlled win rates in Appendix~\ref{sec:win_rate}.

\section{Related Work}
LLMs have shown strong performance in open-ended generation and question answering~\citep{grattafiori2024llama3herdmodels,liu2024deepseekv3,achiam2023gpt,bai2023qwen}, yet their role in fine-grained, instruction-driven revision of academic drafts remains underexplored.

Existing work on LLMs in academic writing typically falls into four categories: (1) \textit{end-to-end generation}, which lacks controllability and author alignment~\citep{porsdam2023autogen,shao2024assisting,jiang2024unknown,weng2024cycleresearcher,asai2024openscholar,schmidgall2025agentlaboratory}; (2) \textit{idea generation}, which raises concerns around authorship and originality~\citep{baek2024researchagent,ghafarollahi2024sciagents,li2024chain,si2024can,gu2024ideagen}; (3) \textit{review assistance and QA systems}, which aid evaluation but do not improve writing quality~\citep{d2024marg,liang2024can,lu2024aiscientist,peerreviewmt2024,chen2024scholarchemqa,lala2023paperqa,song2024cs,lin2024biokgbench}; and (4) \textit{superficial polishing tools}, which perform minor grammatical edits without understanding academic context~\citep{cowriter,TeXGPT}. In contrast, we focus on refining complete drafts with structured, instruction-driven edits that align with the author's intent.

Human--AI collaboration has succeeded across domains such as coding~\citep{chen2021evaluating,nijkamp2022codegen}, medical diagnosis~\citep{chen-etal-2024-towards-injecting}, and peer review~\citep{d2024marg,liang2024can,lu2024aiscientist}. Hybrid workflows, where humans retain control while LLMs provide targeted assistance, are increasingly seen as the most effective model for research~\citep{ashkinaze2024ai,liu2024ai,padmakumardoes}. Although recent work has emphasized controllable generation~\citep{ge2025llmsvulnerable}, few studies address paper revision as a structured, iterative process. Our work fills this gap by leveraging LLMs for instruction-based paragraph revision, helping researchers improve clarity, coherence, and academic rigor while focusing on higher-level reasoning. Due to space constraints, the full discussion of related work is provided in Appendix~\ref{sec:related_work}.

\section{Conclusion}

This paper presents a \textbf{methodological shift} for applying LLMs to academic writing, moving beyond generic text generation towards a controllable, collaborative revision process. Our core contribution is a framework that leverages full-document context and explicit, \textbf{criteria-guided} instructions to enable fine-grained, \textbf{context-aware} paper improvements. By formulating a controllable post-training objective based on this principle, we successfully align model outputs with the rigorous demands of scholarly communication. The empirical success of this approach validates that a focus on controllable, iterative refinement, rather than autonomous generation, is a more trustworthy paradigm for \textbf{human-AI collaboration} in academia.

\section*{Discussion and Broader Implications}
\label{sec:discussion}

\subsection*{Broader Impact}
\label{sec:impact}
The future AI-assisted academic research raises critical concerns.  We analyze these concerns from the perspective of \texttt{XtraGPT}.

\begin{itemize}
    \item \textbf{Potential for Human Researcher Passivity}: One widespread concern is the potential for over-reliance on AI, leading to diminished human effort, creativity, and critical thinking, as AI could handle various stages like idea generation, writing, and reviewing. Our framework, however, adheres to a human-AI collaboration paradigm where the human researcher retains agency and control. Authors are required to have a strong motivation to generate core ideas and initial drafts, reflecting their intellectual investment and desire for recognition. This collaborative process can be viewed as a positive \textit{feedback loop}: the AI's assistance in refining the presentation of core ideas through revisions can, via psychological phenomena such as the \textit{self-fulfilling prophecy}, \textit{reinforce} the human author's motivation and drive for high-quality output. This fosters a virtuous cycle that encourages authors to be more active and engaged in producing and refining their work, rather than becoming passive.

    \item \textbf{Proliferation of Low-Quality Papers and Quantity Inflation}: AI, particularly in uncontrolled end-to-end generation scenarios, poses a risk of enabling the mass production of low-quality or superficially polished papers, potentially inflating publication numbers without commensurate scientific value. In our framework, the AI functions as an assistant specifically for improving existing drafts based on explicit instructions and established writing criteria (such as those informed by academic guides). The initial effort required from the human author to develop high-quality ideas and preliminary drafts is significant and remains a crucial, valuable step that underpins the current positive development of the research community. \texttt{XtraGPT} is designed to help authors present their ``already valuable'' work more effectively and rigorously, implicitly discouraging the dissemination of poorly conceived work and supporting the critical refinement process that characterizes high-quality academic output, thereby helping the community manage article quality rather than promoting quantity over substance, thereby mitigating the sustainability crisis in AI conferences~\cite{chen2025position}.

    \item \textbf{Misalignment with Human Values and Scientific Principles}: Concerns exist that AI might generate content that deviates from human researchers' values, core scientific principles, ethical considerations, or specific conference norms. \texttt{XtraGPT}'s emphasis on controllability and instruction-following is designed to mitigate this risk. The model is trained and operates under constraints that aim to keep it closely aligned with the human author's core intent and the integrity of the original manuscript. In every collaborative interaction, the human author maintains overall judgment and intellectual control, ensuring that the final revised content reflects their will and adheres to academic standards and ethical guidelines, which are implicitly learned from the training data and explicitly guided by user instructions.
\end{itemize}

This work specifically targets the iterative process of paper revision, which is crucial for refining scientific communication, and aims to offer novel insights and tools to the community. We strongly advocate for increased attention to the ability of large models to adhere to core scientific principles and community standards, and emphasize that evaluation metrics for AI in academic assistance should move beyond traditional natural language processing scores to incorporate measures of adherence to these critical norms and principles. This consideration is vital for the responsible development and deployment of AI across all research assistance tasks.

\subsection*{Limitations}
\label{sec:limitations}
We discuss several limitations in the current work. First, the \texttt{ReviseQA} dataset, while comprehensive for its design goals, is constructed from papers in top-tier venues of the AI/ML domain and leverages LLMs for generating instruction--revision pairs. This introduces potential limitations in terms of domain specificity and generator bias. The learned revision strategies may be heavily tailored to the writing style and norms prevalent in these venues, potentially limiting generalizability to other scientific disciplines.

Second, while \texttt{XtraGPT} demonstrates strong performance on instruction-guided paragraph revision, it is an LLM-based system and shares inherent limitations of this technology. The model's ability to maintain long-term context or internal state across multiple, complex iterative revision cycles on a full paper remains an area for further exploration.

Third, evaluating the true impact of fine-grained revisions on overall paper quality is inherently challenging. Our reliance on LLM-as-a-Judge metrics (LC-Win Rate) and even the \textsc{AI-SCIENTIST} tool for full-paper scoring (Appendix~\ref{sec:scorepaper}) faces limitations. LLM judges may not perfectly align with human expert reviewers on all aspects of scientific rigor and novelty. The difficulty of reliably scoring full papers with current AI tools highlights the need for a definitive, automated measure of the impact of \texttt{XtraGPT}'s revisions.

Finally, as discussed in Section \ref{sec:impact}, the potential for over-reliance, introduction of bias from training data, and challenges in interpreting model suggestions are inherent risks associated with LLM-assisted writing tools. While our collaborative design aims to mitigate these, they represent limitations that must be carefully considered during deployment and use.

\subsection*{Potential Risks and Ethical Statement}
\label{sec:ethics}
This work does not advocate the use of LLM to replace human creativity or research ethics standards. Key ethical considerations include Authorship and originality (may blur authorial voice but can be mitigated by human control), bias in training data (ICLR papers, expert annotators), quality and accuracy (may be illusory in scientific contexts and require mandatory human validation), dependence and misuse (may degrade writing skills and superficially polish flawed work), and transparency (limited interpretation of specific recommendations). Our human-computer collaborative design and open source approach aims to promote responsible and transparent use.

\section*{Acknowledgments}
This research is supported by the Ministry of Education AcRF Tier 1 grant (No. T1 251RES2315) in Singapore, Google South \& Southeast Asia Research Award 2025, and the National Research Foundation, Singapore and Infocomm Media Development Authority under its Trust Tech Funding Initiative. Any opinions, findings and conclusions or recommendations expressed in this material are those of the author(s) and do not reflect the views of National Research Foundation, Singapore and Infocomm Media Development Authority.

We also thank the AMD Heterogeneous Accelerated Compute Clusters (HACC) program for the generous hardware donation. We thank Zhen Zhang, Feng Yu, Michael Shieh, Yao Chen, Benyou Wang, Anningzhe Gao, Wentao Ge, Fei Yu, Qi Li, Junying Chen, Zhengyang Cai, and Shunian Chen for their helpful feedback.

\begin{hyphenrules}{nohyphenation}
\setlength{\bibsep}{.5ex plus .8ex}
\bibliographystyle{unsrtnat}
\bibliography{main}
\end{hyphenrules}

\clearpage
\appendix

\section{Analysis of Generated Text Naturalness}
\label{app:ai_detection}

\begin{table*}[h]
\centering
\caption{AI detection metrics on the \texttt{ReviseQA} test set ($N=7,000$). We report results for both Fast-DetectGPT \cite{bao2023fast} and Binoculars \cite{hans2024binoculars}. \texttt{XtraGPT} models (after revision) consistently achieve scores classified as ``Human-written'' (before revision) across both detection methods.}
\label{tab:ai_detection_appendix}
\resizebox{0.95\textwidth}{!}{%
\begin{tabular}{llcccc}
\toprule
\textbf{Method} & \textbf{Metric} & \textbf{Human (Source)} & \textbf{\texttt{XtraGPT}-7B} & \textbf{\texttt{XtraGPT}-14B} & \textbf{Interpretation} \\
\midrule
\multirow{2}{*}{\textbf{Fast-DetectGPT}} 
 & Criterion ($\downarrow$) & $-1.8566$ & $-0.2183$ & $-0.2939$ & Lower is more human-like \\
 & Probability (AI\%) & $27.74\%$ & $38.22\%$ & $37.27\%$ & $< 50\%$ indicates Human \\
\midrule
\multirow{2}{*}{\textbf{Binoculars}} 
 & Score ($\uparrow$) & $1.0143$ & $0.9292$ & $0.9285$ & $> 0.9015$ indicates Human \\
 & Std. Dev & $0.1107$ & $0.0973$ & $0.0966$ & -- \\
\bottomrule
\end{tabular}
}
\end{table*}

To assess whether the revisions generated by \texttt{XtraGPT} exhibit distinguishable machine-generated artifacts, we conducted a large-scale detection analysis on the full \texttt{ReviseQA} test set ($N=7,000$). We employed two state-of-the-art zero-shot metrics that target \textbf{orthogonal statistical artifacts}: \textbf{Binoculars} \cite{hans2024binoculars}, which evaluates global fluency via perplexity ratios, and \textbf{Fast-DetectGPT} \cite{bao2023fast}, which assesses local sampling variance via probability curvature. This combination ensures a robust evaluation of naturalness across different distributional dimensions.

The quantitative results are summarized in Table~\ref{tab:ai_detection_appendix}. We observe that:

\begin{itemize}
    \item \textbf{Evasion of AI Detection:} Both \texttt{XtraGPT}-7B and \texttt{XtraGPT}-14B consistently \textbf{yield scores classified as ``Human-written'' by both detectors}. For Fast-DetectGPT, the AI probabilities are approximately $38\%$, well below the standard decision threshold of $50\%$. Similarly, for Binoculars, the scores are approximately $0.93$, exceeding the detection threshold of $0.9015$ (where higher scores indicate human text).
    
    \item \textbf{High Naturalness:} These results indicate that \texttt{XtraGPT} generates highly naturalistic academic text that aligns closely with the linguistic patterns of human writing. The model successfully avoids common statistical fingerprints that typically trigger these detectors in other domains.
    
    \item \textbf{Domain Characteristics:} We note that while a statistical gap remains between the original human text (Human Source) and the model outputs (e.g., Human Criterion $-1.86$ vs. Model $-0.22$), the model outputs fall safely within the distribution of authentic-sounding text. This suggests that \texttt{XtraGPT} preserves the high-quality, low-entropy nature characteristic of academic writing.
\end{itemize}
\section{Related Work}
\label{sec:related_work}

\paragraph{LLMs Assist Academic Writing}

Research on LLMs for academic writing falls into four primary categories. First, \textit{automated paper generation} attempts to produce complete papers but often lacks user control and academic rigor \citep{shao2024assisting, jiang2024unknown, weng2024cycleresearcher, asai2024openscholar, schmidgall2025agentlaboratory}. Second, \textit{research ideation} employs LLMs to propose novel ideas and methodologies, though concerns regarding authorship and originality persist \citep{baek2024researchagent, ghafarollahi2024sciagents, li2024chain, si2024can}. Third, thanks to the success of retrieval by instruction \citep{sun2024mair}, \textit{automated reviewing and research question answering} assist in literature searches and manuscript evaluations but do not directly refine writing quality \citep{d2024marg, liang2024can, lu2024aiscientist, peerreviewmt2024, asai2024openscholar, chen2024scholarchemqa, lala2023paperqa, song2024cs, lin2024biokgbench}. Lastly, \textit{LLM-assisted writing tools} enhance grammar and style and \cite{shi2023relatedwork} improves a small paragraph of paper, they lack deep contextual awareness necessary for high-quality academic discourse \citep{cowriter,TeXGPT}.

\paragraph{LLMs Assist Research}

Beyond writing, LLMs are increasingly utilized in autonomous research \cite{weng2024cycleresearcher,lyu2026evoscientist}. \cite{swanson2024virtual} introduced LLM agents functioning as research assistants, integrating human feedback into scientific workflows. ChemCrow \citep{m2024augmenting} and Coscientist \citep{boiko2023autonomous} highlight LLM-led ideation and experimentation in chemistry, while ResearchAgent \citep{baek2024researchagent} automates research generation, iterative refinement, and review. AI Scientist \citep{lu2024aiscientist} extends automation to coding, experimentation, and manuscript review. Despite these advancements, studies caution that LLMs require human oversight to ensure reproducibility and scientific rigor \citep{si2024can}.

\paragraph{Gaps and Contributions}

LLMs also contribute to research tasks such as code generation \citep{chen2021evaluating, nijkamp2022codegen}, literature search \citep{ajith2024litsearch, kang2024researcharena, press2024citeme, li2024scilitllm}, and automated paper reviewing \citep{d2024marg, liang2024can, lu2024aiscientist, weng2024cycleresearcher}. While they support ideation \citep{si2024can}, concerns about reduced creativity and homogenization persist \citep{chakrabarty2024art, anderson2024homogenization}. Hybrid human-LLM approaches are seen as the most effective way to enhance research workflows \citep{ashkinaze2024ai, liu2024ai, padmakumardoes}.

Recently, the controllable generation of LLMs have been emphasized \citep{ge2025llmsvulnerable}. While much work has focused on using LLMs for idea generation, review, and automation, little research directly addresses refining research papers to enhance coherence, clarity, and adherence to academic standards. Our work bridges this gap by leveraging LLMs specifically for structured refinement, allowing researchers to focus on deeper reasoning tasks while ensuring scholarly rigor.

\paragraph{LLM simulation}
Researchers have increasingly utilized Large LLMs to construct simulations, treating LLM agents as proxies for humans to perform actions and interactions \citep{Park2023GenerativeAgents, lin2023agentsims, kong2024platolm, wang2024megaagent}. These simulations have shown promise in diverse fields such as society, economics, policy, and psychology \citep{Park2023GenerativeAgents, li2024cryptotrade, chen2024agentcourt}, while also serving as data generators and evaluators for LLM training \citep{tang2024synthesizing, zhang2024regurgitative}. However, LLMs face significant limitations in simulation tasks. Studies \citep{ai2024cognition, petrov2024limited, hu2024quantifying, lee2024language, tang2026llm} highlight their inability to maintain contextual consistency and produce fine-grained outputs. For example, \citet{lee2024language} found that LLMs exhibit consistent values and preferences even when role-playing diverse personas, underscoring their lack of adaptability and nuanced understanding.

\section{Prompts}
Figure \ref{fig:prompt4score} shows the prompt for QA.

\begin{figure}[h]
\centering
\begin{tcolorbox}[colback=blue!5!white, colframe=blue!70!black, title=The Prompt for QA, width=0.9\textwidth, sharp corners=south]

Act as an expert model for improving articles **PAPER\_CONTENT**. 

\texttt{<SELECTED\_CONTENT>}

\texttt{User Selected}

\texttt{</SELECTED\_CONTENT>}

\texttt{<QUESTION>}

\texttt{<User Question>}

\texttt{</QUESTION>}

\end{tcolorbox}
\caption{Prompt for QA}
\label{fig:prompt4score}
\end{figure}

\begin{figure}[h!]
\centering
\begin{tcolorbox}[colback=blue!5!white, colframe=blue!70!black, title=The Prompt for Generating QA pairs, width=0.98\textwidth, sharp corners=south,boxsep=1pt, top=2pt, bottom=2pt, left=3pt, right=3pt]

You are an advanced language model designed to assist users in improving their articles. Users will provide an article in LaTeX or Markdown format and specify a **section** along with **criteria** for improvement. Your task is to identify a specific selected content from the provided section, align it with the given criteria, and offer actionable feedback to improve the content.

Instructions:

1. **Role 1**: We have a paper improvement task with a specific criteria '{criteria['prompt']}'. Now play a role as an author of the provided paper content. Select a specific content from the section '{section}' (or equivalent), and ask a chatbot assistant to help you improve that selected content.%

   - **The selected paper content must be a worth-improving paragraph(s)** that might not achieve the standards of the criteria '{criteria['prompt']}', and that content should come from the section '{section}'. The selected content will be labeled as **BEFORE IMPROVEMENT**.

   - Provide a concise, conversational improvement-related question labeled as **QUESTIONS**. These questions should not explicitly tell what rules or standards to follow or what the specific goal should be. Instead, offer a high-level instruction that may hint at the criteria without stating them directly. The aim is to allow for creativity and subtle alignment with the criteria.

   - Keep the question short and conversational.

2. **Role 2**: Act as an expert model for improving articles.

The revised version of the selected content should be labeled as AFTER IMPROVEMENT and specifically address the QUESTIONS on BEFORE IMPROVEMENT above. Avoid adding unnecessary length, unrelated details, overclaims, or vague statements. Focus on clear, concise, and evidence-based improvements that align with the overall context of the paper.

Provide a detailed explanation of the changes made, labeled as EXPLANATION, with clear references to the paper's content. Ensure the explanation demonstrates how the revisions align with the context and criteria of the paper.

--- PAPER CONTEXT STARTS

{paper\_latex}

--- PAPER CONTEXT ENDS

Response Format (must be strictly followed):

--- BEFORE IMPROVEMENT STARTS

\texttt{<}Selected content\texttt{>}

--- BEFORE IMPROVEMENT ENDS

--- QUESTIONS START  

\texttt{<}Concise, improvement-related question based on the criteria '{criteria['prompt']}'\texttt{>}

--- QUESTIONS END  

--- AFTER IMPROVEMENT STARTS  

\texttt{<}Revised version of the selected content to answer the **Questions** above\texttt{>}
--- AFTER IMPROVEMENT ENDS  

--- EXPLANATION STARTS  

\texttt{<}An explanation of the changes made, showing how they align with the context of the article and address the criteria. Include references from the paper context where relevant.\texttt{>}

--- EXPLANATION ENDS

\end{tcolorbox}
\vspace{-4mm}
\caption{Prompts for Generating \texttt{ReviseQA}}
\label{prompt4GENQA}
\end{figure}

\begin{figure}[h!]
\centering
\begin{tcolorbox}[colback=blue!5!white, colframe=blue!70!black, title=The Prompt for Judging, width=0.9\textwidth, sharp corners=south]
You are a highly efficient assistant, who evaluates and rank large language models (LLMs) based on the quality of their responses to given prompts. This process will create a leaderboard reflecting the most accurate and human-preferred answers.

I require a leaderboard for various large language models. I'll provide you with prompts given to these models and their corresponding responses. Your task is to assess these responses, ranking the models in order of preference from a human perspective. Once ranked, please output the results in a structured JSON format for the make\_partial\_leaderboard function.

Prompt
\begin{verbatim}
{
    "instruction": "{instruction}",
}

\end{verbatim}

Model Outputs

Here are the unordered outputs from the models. Each output is associated with a specific model, identified by a unique model identifier.
\begin{verbatim}
{
    {
        "model": "m",
        "output": "{output\_1}"
    },
    {
        "model": "M",
        "output": "{output\_2}"
    }
}
\end{verbatim}

Task

Evaluate based on the quality and relevance to the instructions. The following is the definition of the quality on the section \texttt{<section>}: \texttt{<criteria["prompt"]>}. If the model's output refers to information beyond \texttt{<Selected content>}, it receives a slightly higher score.
\end{tcolorbox}
\caption{Prompts for Judging (modified from alpaca\_eval\_gpt4\_turbo\_fn).}
\label{prompt4judging}
\end{figure}

\begin{figure}[h!]
\centering
\begin{tcolorbox}[colback=blue!5!white, colframe=blue!70!black, title=The Prompt for Ranking, width=0.9\textwidth, sharp corners=south]

Human: I want you to create a leaderboard of different large-language models. To do so, I will give you the instructions (prompts) given to the models, and the responses of two models. Please rank the models based on which responses would be preferred by humans. All inputs and outputs should be Python dictionaries.

Here is the prompt:
\begin{verbatim}
{
    "instruction": "{instruction}",
}
\end{verbatim}

Here are the outputs of the models:
\begin{verbatim}
{
    "model": "model_1",
    "answer": "{output_1}"
},
{
    "model": "model_2",
    "answer": "{output_2}"
}
\end{verbatim}

Now please rank the models by the quality of their answers, so that the model with rank 1 has the best output. Then return a list of the model names and ranks, i.e., produce the following output:
\begin{verbatim}
[
    {'model': \texttt{<model-name>}, 
    'rank': \texttt{<model-rank>}},
    {'model': \texttt{<model-name>}, 
    'rank': \texttt{<model-rank>}}
]
\end{verbatim}

Your response must be a valid Python dictionary and should contain nothing else because we will directly execute it in Python. Please provide the ranking that the majority of humans would give.
\end{tcolorbox}
\caption{Prompts for Scoring.}
\label{prompt4scoring}

\end{figure}

\section{ICLR 2024 Token Distribution}
\label{sec:tokendistribution}
\begin{figure}
    \centering
    \includegraphics[width=0.6\linewidth]{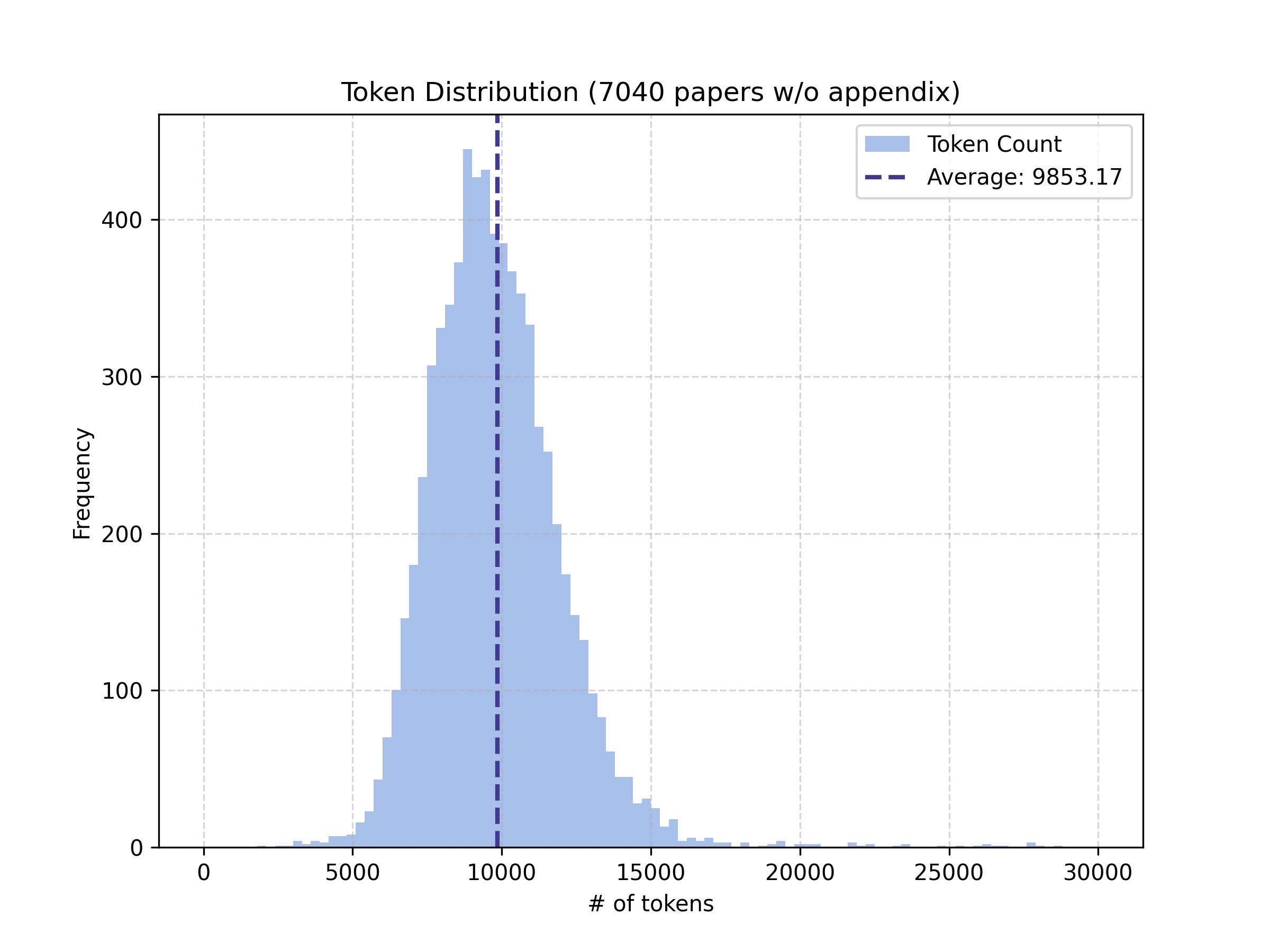}
    \caption{ICLR 2024 Paper Token Distribution (without Appendix)}
    \label{fig:tokendistribution}
\end{figure}

ICLR 2024 token distribution (derived from markdown) is showed in Figure \ref{fig:tokendistribution}.
\section{Human Label Details}
Table \ref{tab:baseline_and_human_alignment_detail},\ref{tab:baseline_and_human_alignment_detail2} shows the details of calculation data on human evaluation.

\begin{table}[h!]
\footnotesize
\centering

\setlength{\tabcolsep}{3pt}
\centering
\begin{tabular}{lccc}
\toprule
QA Controllability Assurance & \textbf{Judge 1} & \textbf{Judge 2} & \textbf{Judge 3}  \\ \midrule
GPT-4o-Mini 
-Instruction Following  & (78+77+72+78+78)/5 & (76+68+79)/3 & (75+81+80+78+73)/5 \\ 
-Criteria Following  & (79+74+63+77+75)/5 & (77+68+79)/3 & (76+81+77+74+75)/5\\ 
-In-Context Ability  & (73+53+48+62+61)/5 & (67+57+75)/3 & (69+76+74+73+72)/5\\ 
-Agree revision?  & (48+48+44+53+53)/5 & (65+56+64)/3 & (67+74+74+72+71)/5 \\
\\
GPT-o1-mini 
-Instruction Following & (79+71+76+76+80)/5 & (79+81+77)/3 & (75+80+78+77+64)/5  \\ 
-Criteria Following & (72+70+74+74+80)/5 & (79+77+75)/3 & (74+80+78+76+63)/5  \\ 
-In-Context Ability& (74+53+58+65+60)/5 & (68+68+68)/3 & (73+75+81+75+64)/5  \\ 
-Agree revision? & (58+50+53+59+60)/5 & (66+66+67)/3 & (72+76+78+76+62)/5  \\\midrule

\end{tabular}
\caption{Human evaluation on improvement acceptance rates before and after paragraph. we ask 3 human evaluators based on 5,3,5 paper, about 100,60,100 questions in score 1-5.
The \textbf{Aggregated} column aggregates the results of 3 human evaluators. }
\label{tab:baseline_and_human_alignment_detail}

\end{table}

\begin{table*}[h!]
\footnotesize
\centering

\setlength{\tabcolsep}{3pt}
\centering
\begin{tabular}{p{3cm}ccc}
\toprule
\protect\includegraphics[width=1.8cm]{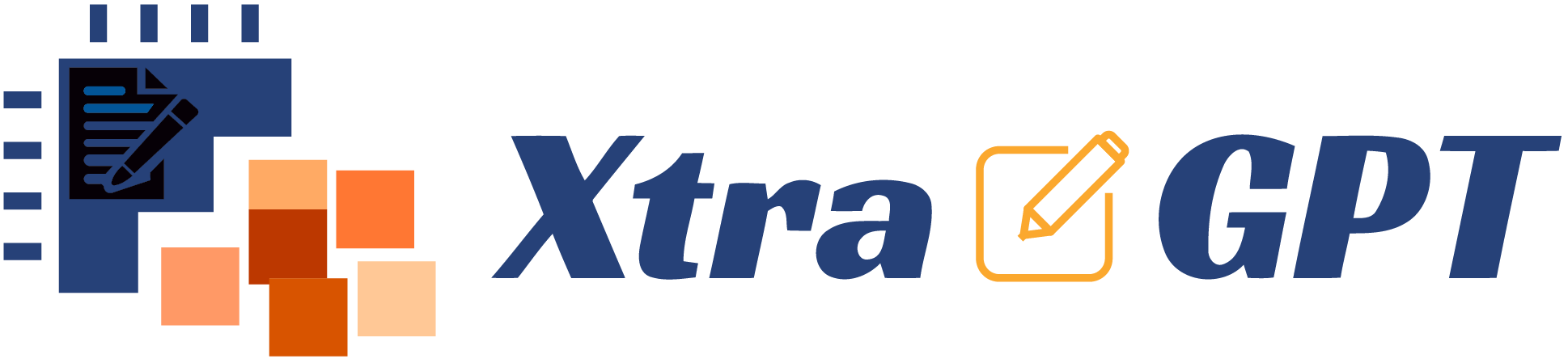} & \textbf{Judge 1} & \textbf{Judge 2} & \textbf{Judge 3} \\ \midrule

Instruction Following & (62+61+72+76+74)/5 & (80+80+79)/3 & (86+80+83+82+78)/5\\
Criteria Following & (60+60+69+72+73)/5 & (74+74)/3 & (82+82+82+81+82)/5  \\
In-Context Ability  & (58+51+48+61+61)5 & (67+69)/3 & (85+80+82+79+80)/5  \\
Agree revision?  & (50+45+44+55+52)/5 & (65+64)/3 & (83+79+82+80+77)/5\\ \hline

\end{tabular}
\caption{\texttt{XtraGPT} Human Evaluation}
\label{tab:baseline_and_human_alignment_detail2}

\end{table*}

\section{Section-Level Criteria Details}
\label{sec:criteria_details}

Section-level criterias are detailed in Figure \ref{fig:criteria_title},\ref{fig:criteria_abstract},\ref{fig:criteria_introduction},\ref{fig:criteria_background},\ref{fig:criteria_evaluation},\ref{fig:criteria_conclusion}.
\begin{figure*}[!h]
\centering
\begin{tcolorbox}[colback=blue!5!white, colframe=blue!70!black, title=Criteria Details of Section Title, width=1.0\textwidth, sharp corners=south]
    \textbf{1. Consistency and Alignment of Title with Paper’s Content:} \\
    Evaluate the degree to which the paper’s title accurately captures its principal topics, arguments, or findings. Does the title reflect the scope and focus of the paper, and is it consistent with the main concepts and keywords presented in the abstract and introduction? Identify any discrepancies or misalignment between the title and the content. \\

    \vspace{5pt}

    \textbf{2. Conciseness and Clarity of Title:} \\
    Evaluate the paper’s title for redundancy. Are there repeated words or concepts that could be removed without changing the core meaning? Does the final title remain succinct, clear, and accurately convey the paper’s main focus or contribution? 
\end{tcolorbox}
\caption{Criteria Details of Section Title}
\label{fig:criteria_title} 
\end{figure*}

\begin{figure*}[!h]
\centering
\begin{tcolorbox}[colback=blue!5!white, colframe=blue!70!black, title=Criteria Details of Section Abstract, width=1.0\textwidth, sharp corners=south]
    \textbf{1. Clarity and Impact of the Conclusion:} \\
    Evaluate the clarity and impact of the conclusion in the abstract. Does it clearly summarize the research steps, highlight key outcomes, and explain the significance of these outcomes for the field of computer science? Are the primary technical advancements and their contributions presented in a concise and unambiguous manner? \\

    \vspace{5pt}

    \textbf{2. Motivation and Purpose in the Abstract:} \\
    Evaluate how well the abstract communicates the research's motivation. Does it clearly articulate the broader issue, concept, or problem in Computer Science that the work addresses? Does it explicitly state the specific research problem being solved and why it is important? \\

    \vspace{5pt}

    \textbf{3. Explanation of Existing Solutions and Research Gap:} \\
    Assess how well the abstract explains the shortcomings of current solutions and highlights the corresponding research gap. Does it clearly articulate why existing methods are insufficient and how the proposed approach addresses these limitations? Is the explanation comprehensible to a wide audience, from domain experts to non-specialists? \\

    \vspace{5pt}

    \textbf{4. Clarity and Adequacy of Proposed Solutions:} \\
    Assess how effectively the abstract communicates the proposed solutions. Does it clearly identify the research gap or problem being addressed, and explain how the proposed solution tackles this gap? Does it highlight the novelty or contribution of the solution, demonstrating its relevance or improvement over existing work? Rate the clarity, completeness, and significance of the explanation provided in the abstract. 
\end{tcolorbox}
\caption{Criteria Details of Section Abstract}
\label{fig:criteria_abstract} 
\end{figure*}

\begin{figure*}[h]
\centering
\begin{tcolorbox}[colback=blue!5!white, colframe=blue!70!black, title=Criteria Details of Section Introduction, width=1.0\textwidth, sharp corners=south]

    \textbf{1. Strength and Clarity of Motivation in the Introduction:} \\
    Evaluate whether the motivation in the Introduction is specific and convincing. Does the paper avoid over-generalization and clearly articulate the significance of the issue? Are concrete examples, statistics, or contextual details used to establish why the problem matters? \\

\vspace{5pt}

    \textbf{2. Review of Existing Approaches in Introduction:} \\
    Assess the thoroughness and clarity of the literature review within the introduction. Does the paper cite and critique relevant prior works, highlighting both their methodologies and limitations? Does the introduction establish how the proposed work builds upon or differentiates itself from existing research, and is there sufficient context provided to demonstrate the significance of the current study? Are any quantitative or qualitative comparisons made where appropriate? \\

\vspace{5pt}

    \textbf{3. Audience Alignment and Appropriateness:} \\
    Evaluate whether the introduction is effectively tailored to its target audience. Is the complexity, depth, and choice of terminology suitable for the presumed background knowledge of the readership? Does the introduction provide sufficient context without oversimplifying or overwhelming the intended audience? \\

\vspace{5pt}

    \textbf{4. Clarity and Visibility of Contributions:} \\
    Assess the clarity and visibility of the paper’s contributions. Are the core contributions explicitly stated in a dedicated paragraph or section toward the end of the introduction? Are they understandable to a broad scientific audience, presented succinctly, and positioned logically following the problem statement and background information? \\

\vspace{5pt}

    \textbf{5. Clarity and Specificity of Problem Definition:} \\
    Evaluate the paper’s problem definition in terms of four key elements: current situation, ideal situation, the gap between them, and how the research aims to address this gap. Are these components clearly stated, distinct, and directly tied to the research objectives? Does the definition provide sufficient clarity and focus for the research? \\

\vspace{5pt}

    \textbf{6. Integration of State-of-the-Art in Problem Framing:} \\
    Evaluate how effectively the introduction incorporates the State-of-the-Art (SOTA) to frame the research problem. Does it include explicit references to key works, methodologies, or findings that highlight relevant gaps or limitations in the field? Is there a clear logical link between the SOTA discussion and the stated research objectives, demonstrating how the proposed work builds upon or extends existing research? 
\end{tcolorbox}
\caption{Criteria Details of Section Introduction}
\label{fig:criteria_introduction} 
\end{figure*}

\begin{figure*}[!h]
\centering
\begin{tcolorbox}[colback=blue!5!white, colframe=blue!70!black, title=Criteria Details of Section Background, width=1.0\textwidth, sharp corners=south]
    \textbf{1. Contextual Relevance and Clarity of Background:} \\
    Assess how effectively the background section establishes context for the research. Does it provide a clear overview of the broader field in computer science, then narrow down to the specific problem? Does the paper avoid making unwarranted assumptions about the reader’s prior knowledge? Finally, does it clarify why addressing the problem is important to the field? \\

    \vspace{5pt}

    \textbf{2. Coverage of Key Preliminary Concepts:} \\
    Evaluate the thoroughness and clarity of the paper’s background or preliminary section. Does it introduce and define all the critical concepts, algorithms, or theorems necessary to understand the technical contributions? Are these concepts clearly explained, logically organized, and accessible to readers who are not experts in the field? Does the paper use consistent terminology and adequately explain symbols, abbreviations, or specialized terms before their first usage? \\

    \vspace{5pt}

    \textbf{3. Clarity and Consistency of Terminology:} \\
    Assess the clarity and consistency of the key terms introduced in the background section. Are all critical terminologies defined at their first occurrence and used consistently throughout the paper? Does the paper avoid undefined shifts or redefinitions of terms, and does it align terminology with standard conventions in the field? 
\end{tcolorbox}
\caption{Criteria Details of Section Background}
\label{fig:criteria_background} 
\end{figure*}

\begin{figure*}[!h]
\centering
\begin{tcolorbox}[colback=blue!5!white, colframe=blue!70!black, title=Criteria Details of Section Evaluation, width=1.0\textwidth, sharp corners=south]
    \textbf{1. Experimental Setup Clarity and Reproducibility:} \\
    Evaluate how clearly and thoroughly the experimental setup is described. Does the paper provide all necessary information on hardware/software configurations, parameter settings, data preprocessing steps, and any contingency procedures, such that others could replicate the experiments with the same resources? \\

    \vspace{5pt}

    \textbf{2. Depth and Clarity of Figures and Tables Analysis:} \\
    Evaluate the thoroughness and clarity of the paper’s analysis of figures and tables. Are the data clearly explained and linked to the research objectives or hypotheses? Do the authors discuss trends, patterns, or anomalies, and interpret quantitative metrics in a way that highlights their significance? Is there a clear comparison to baselines or related work, demonstrating how the results fit into or advance the field? Do the authors emphasize key takeaways and practical or theoretical implications arising from the findings? \\

    \vspace{5pt}

    \textbf{3. Experimental Support for Main Innovations:} \\
    Evaluate how thoroughly the paper’s main innovations or contributions are backed by experimental evidence. Does the paper provide direct tests or comparisons to validate each innovation? Are quantitative or qualitative results clearly linked to the claims made, with appropriate metrics and comparisons against baselines or existing methods? Are ablation studies or sensitivity analyses included to demonstrate the significance of each component? If certain claims are not experimentally supported, have the authors either provided additional experiments or adjusted their claims accordingly? 

\end{tcolorbox}
\caption{Criteria Details of Section Evaluation}
\label{fig:criteria_evaluation} 
\end{figure*}

\begin{figure*}[!h]
\centering
\begin{tcolorbox}[colback=blue!5!white, colframe=blue!70!black, title=Criteria Details of Section Conclusion, width=1.0\textwidth, sharp corners=south]
    \textbf{1. Broader Impact and Future Directions:} \\
    Assess the thoroughness of the paper's conclusion or discussion sections in addressing the broader impact of the research. Does the paper provide specific and clear avenues for future work? \\
    
    \vspace{5pt}
    
    \textbf{2. Clarity and Impact of Key Innovations and Findings:} \\
    Evaluate whether the conclusion effectively highlights the paper’s key innovations. 
\end{tcolorbox}
\caption{Criteria Details of Section Conclusion}
\label{fig:criteria_conclusion} 
\end{figure*}
\section{Hyperparams}
\label{sec:hypers}
\begin{table}[H]
\centering
\resizebox{0.45\textwidth}{!}{
\begin{tabular}{lc}
\toprule
\textbf{Hyperparameter} & \textbf{value} \\
\midrule
Batch Size & \{1,2\} \\
Cut-off Len & 16384\\
max\_new\_tokens & 512\\
Epoch & \{10,20\}\\
Learning Rate & \{1e-5,2e-5\} \\
\bottomrule
\end{tabular}

}

\caption{\label{table:hparams_train}
Hyperparameters
}
\end{table}

\section{LoRA vs. Full SFT}
\label{sec:lora_tuned}
\begin{table*}[h]
\footnotesize
\centering
\caption{Length-controlled (LC) win rates.}
\label{tab:lora_win_rate}
\resizebox{\textwidth}{!}{
\setlength{\tabcolsep}{3pt}
\begin{tabular}{lcccccccc}
\toprule
\textbf{Models} & \textbf{Title} & \textbf{Abstract} & \textbf{Introduction} & \textbf{Background} & \textbf{Evaluation} & \textbf{Conclusion} & \textbf{Overall}$\uparrow$\\ 
\midrule
\rowcolor{lightgray}
\multicolumn{8}{c}{\texttt{XtraGPT}-7B; base: \texttt{Qwen-2.5-7B-Instruct}(\textit{anchor}$\uparrow$)} \\
% Qwen-2.5-7B-Instruct & 50.41 & 47.11 & 43.71 & 46.56 & 46.05 & 49.79 & 46.44 \\
% Qwen-2.5-7B-Instruct (10\%) & 50.48 & 38.44 & 42.06 & 53.33 & 46.82 & * & * \\
% Qwen2.5-7B-Instruct-paperfull214/checkpoint-1000-new & 51.85 & 50.32 & 44.85 & 44.44 & 46.84 & 50.00 & * \\
Qwen2.5-7B-Instruct Lora Tuned & 44.58 & 45.98 & 47.01 & 40.75 & 35.21 & 32.27 & 42.17 \\ 
Qwen-2.5-7B-Instruct (Base) & 39.93 & 45.14 & 45.64 & 39.28 & 33.87 & 31.17 & 40.80 \\
\bottomrule
\end{tabular}}
\end{table*}
We investigate LoRA~\cite{hu2022lora} as a parameter-efficient alternative (1 epoch, learning rate $2.0 \times 10^{-5}$). As shown in Table~\ref{tab:lora_win_rate}, both the LoRA variant (42.17) and the Base model (40.80) are consistently outperformed by our full SFT anchor. The performance gap, especially in the Evaluation and Conclusion sections, suggests that full-parameter fine-tuning is more effective at capturing the complex, long-context dependencies required for high-quality academic revision than the lower-rank adaptation.

\section{Win Rate}
\label{sec:win_rate}
Table \ref{tab:win_rate} shows the win rate without length control, which is unreasonable compared to Table \ref{tab:lc_win_rate}.

\begin{table*}[t]
\footnotesize
\centering
\caption{Average output length of each model on the 7,000-instance \textsc{ReviseQA} test set. Some models tend to produce significantly longer responses, occasionally including unsolicited explanations.}
\label{tab:model_output_length}
\resizebox{\textwidth}{!}{
\setlength{\tabcolsep}{3pt}
\begin{tabular}{lcccccccc}
\toprule
\textbf{Models} & \textbf{Title} & \textbf{Abstract} & \textbf{Introduction} & \textbf{Background} & \textbf{Evaluation} & \textbf{Conclusion} \\ 
\midrule
Original Text & 14.5 & 174.9 & 200.1 & 238.9 & 170.8 & 151.3 \\
% Qwen2-72B-Instruct & 84.1 & 210.6 & 280.9 & 332.0 & 403.9 & 242.5 \\
Phi-4 & 126.8 & 265.2 & 339.5 & 357.1 & 348.3 & 329.6 \\
DeepSeek-v3-671B & 17.5 & 158.7 & 243.5 & 258.8 & 245.7 & 217.4 \\
GPT-4o-Mini & 16.9 & 189.4 & 254.4 & 265.8 & 227.9 & 210.9 \\
% \rowcolor{lightgray}
% \multicolumn{8}{c}{\protect\includegraphics[width=1.7cm]{XtraGPT_221.png} (\textit{anchor}$\uparrow$)} \\
Qwen-2.5-7B-Instruct & 21.2 & 187.3 & 272.2 & 258.4 & 264.1 & 242.9 \\
% Qwen-2.5-7B-Instruct & 51.34 & 46.94 & 46.47 & 43.46 & 46.63 & 44.91 & 46.47 \\
% \textbf{DeepSeek-R1-Distill-Llama3-8B}\\
% \textbf{qwen-1.5-110B}\\
% Qwen-2.5-7B-Instruct & 46.18 & 63.43 & 45.74 & 46.90 & 46.44 & 49.50 & \textcolor{red}{49.98} & 49.70 \\
Qwen-QWQ-32B & 40.7 & 224.1 & 272.9 & 265.4 & 308.7 & 245.5 \\
% DeepSeek-R1-Distill-Llama-8B & 97.9 & 171.0 & 207.9 & 252.1 & 269.5 & 190.3 \\
% Phi-3.5-mini-instruct (3.8B) & 30.60 & 54.36 & & 37.01 &38.03 &56.71&\\
Llama-3.2-3B-Instruct & 327.4 & 445.2 & 464.4 & 479.3 & 472.3 & 443.7 \\
Qwen2.5-1.5B-Instruct & 89.5 & 291.8 & 316.2 & 336.8 & 361.4 & 352.9 & \\
Qwen2.5-72B-Instruct & 16.2 & 195.1 & 275.6 & 284.2 & 305.9 & 246.3 & \\
XtraGPT-14B & 15.9 & 173.8 & 231.0 & 251.6 & 242.3 & 205.5 & \\
XtraGPT-7B & 15.5 & 180.2 & 233.1 & 251.2 & 250.1 & 213.5 & \\
XtraGPT-3B & 15.9 & 179.7 & 237.8 & 254.2 & 251.8 & 214.8 & \\
XtraGPT-1.5B & 15.7 & 183.0 & 232.4 & 248.9 & 251.6 & 213.1 \\
GPT-3.5-Turbo & 57.9 & 144.0 & 178.3 & 189.8 & 169.2 & 146.4 \\
DeepSeek-R1-Distill-Qwen-7B & 419.1 & 668.5 & 890.3 & 942.3 & 1081.2 & 780.7 \\
% Llama-3.1-8B-Instruct & 139.5 & 288.7 & 400.7 & 422.0 & 385.6 & 204.7 \\
\bottomrule
\end{tabular}}
\end{table*}

\begin{table*}[h]
	\footnotesize
	\centering
	\resizebox{\textwidth}{!}{
		\setlength{\tabcolsep}{3pt}
		\begin{tabular}{lccccccc}
			\toprule
			\textbf{Models}                            & \textbf{Title} & \textbf{Abstract} & \textbf{Introduction} & \textbf{Background} & \textbf{Evaluation} & \textbf{Conclusion} & \textbf{Average}$\uparrow$ \\
			\midrule
			Qwen2-72B-Instruct                         & 53.57          & 70.93             & 77.52                 & 86.76               & 91.90               & 73.71               & 75.73                      \\
			GPT-4o-Mini                                & 65.57          & 59.71             & 70.05                 & 67.81               & 70.86               & 62.14               & 66.02                      \\
			Qwen-QWQ-32B-Preview                       & 62.97          & 66.42             & 61.33                 & 73.24               & 72.48               & 74.29               & 69.88                      \\
                Deepseek-V3-671B & 63.79           & 59.29              & 66.19                  & 61.24                & 88.95                & 58.57                & 66.33
            \\
			Qwen-2.5-7B-Instruct                       & 60.79          & 70.93             & 60.52                 & 56.48               & 74.48               & 70.43               & 65.60                      \\
			\rowcolor{lightgray}
			\multicolumn{8}{c}{\textbf{XtraGPT}-7B (base: \texttt{Qwen-2.5-7b-instruct}) (\textit{anchor}$\uparrow$)}                                                                                                                \\
			Llama-3.1-8B-Instruct                      & 47.41          & 39.64             & 41.24                 & 55.24               & 55.71               & 30.29               & 44.92                      \\
			Qwen2.5-1.5B-Instruct                      & 34.36          & 32.39             & 26.14                 & 21.24               & 26.48               & 31.29               & 28.65                      \\
			GPT-3.5-Turbo                              & 28.57          & 20.79             & 19.38                 & 20.95               & 23.05               & 11.43               & 20.70                      \\
			Llama-3.2-3B-Instruct                      & 27.43          & 9.29              & 10.90                 & 9.71                & 14.67               & 6.43                & 13.07                      \\
			\bottomrule
		\end{tabular}}
	\caption{Win rates of various models against \textsc{XtraGPT} (\textit{anchor}) across different evaluation categories. Models are ranked in descending order based on their averaged win rates.}
	\label{tab:win_rate}
\end{table*}
\section{Annotators for Controllable Quality Assurance}
Figure \ref{fig:criterias} show annotators for controllable quality assurance.

\begin{figure*}[t]
\vspace{-10mm}
\begin{AIbox}{Criteria}

Each QA pair is evaluated based on four metrics, each scored from 1 to 5:

Evaluation Metrics (1-5 Scoring Criteria)

1. \textbf{Instruction Following: }
Evaluate whether the answer correctly follows the given instruction.

1 – The answer completely ignores or contradicts the instruction.

2 – The answer only partially follows the instruction, with major missing elements.

3 – The answer follows the instruction but lacks completeness or clarity.

4 – The answer mostly follows the instruction with minor inconsistencies.

5 – The answer strictly follows and fully satisfies the instruction.

2. \textbf{Criteria Following: }
Evaluate whether the revised text improves the original content based on predefined criteria.

1 – The revision does not follow any criteria and worsens the content.

2 – The revision attempts to follow the criteria but makes the content unclear.

3 – The revision follows the criteria but does not provide a significant improvement.

4 – The revision improves clarity and correctness while adhering to the criteria.

5 – The revision strictly follows the criteria and significantly improves the original content.

3. \textbf{In-Context Ability:}
Evaluate whether the model’s output appropriately references information within Selected Content.

1 – The output ignores Selected Content and adds irrelevant external information.

2 – The output relies on external information without justification.

3 – The output primarily references Selected Content but includes minor unrelated details.

4 – The output correctly refers to Selected Content with minimal external additions.

5 – The output strictly remains within Selected Content while providing a relevant and precise response.

4. \textbf{Agree Revision:}
Evaluate whether the revision is convincing enough for the user to adopt it as a replacement.

1 – The revision is clearly worse than the original text.

2 – The revision is slightly better but has major flaws, making it unlikely to be adopted.

3 – The revision is neutral or slightly better, but adoption is uncertain.

4 – The revision is clearly better, and most users would likely adopt it.

5 – The revision is significantly better, and users would confidently adopt it.

\end{AIbox} 
\vspace{-3mm}
\caption{The criteria for human instructors.}
\vspace{-3mm}
\label{fig:criterias} 
\end{figure*}

\begin{figure*}
    \centering
    \includegraphics[width=0.98\linewidth]{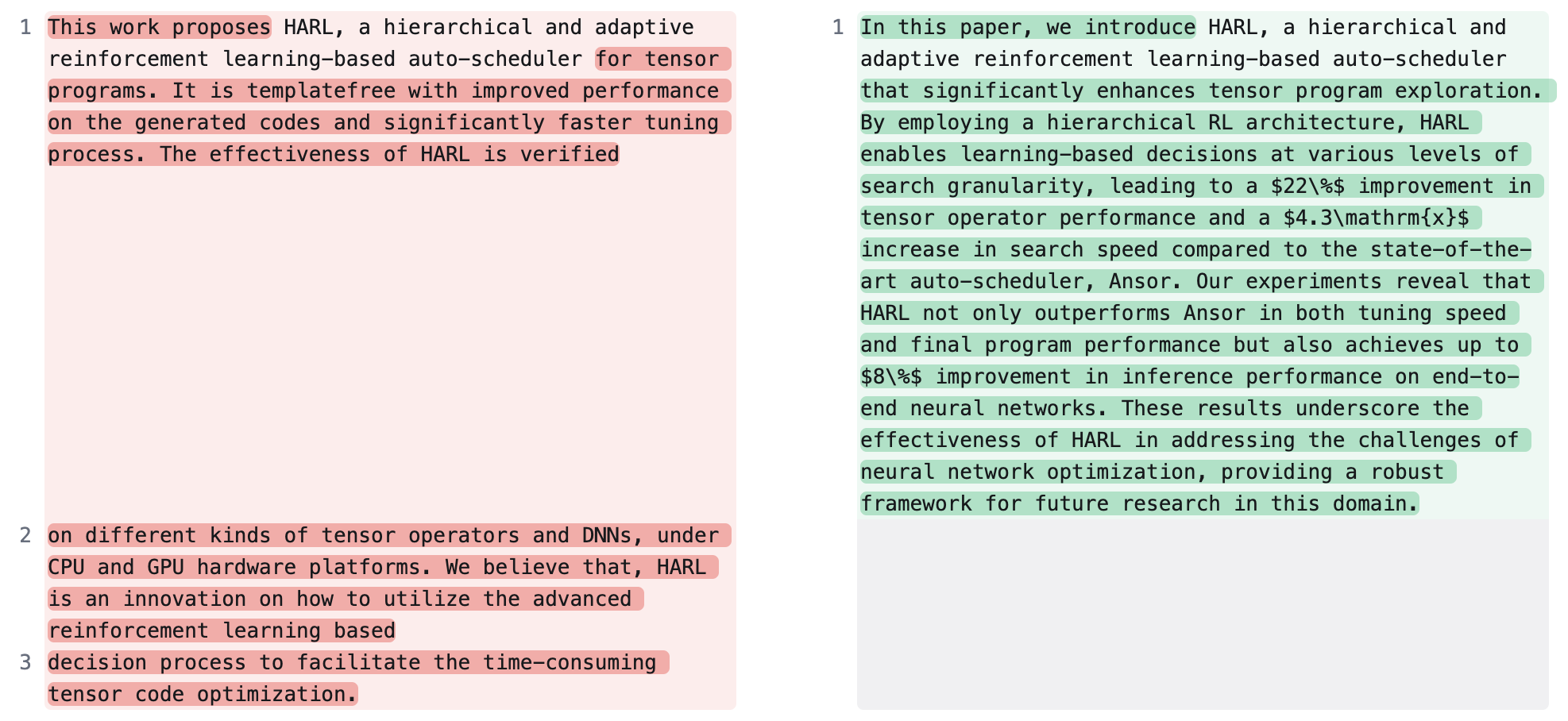}
    \caption{A comparison of the paragraph before and after revision.}
    \label{fig:ui2}
\end{figure*}

\section{Case Study}
We chose HARL~\citep{harl} in Figure \ref{fig:harl} as a case study to demonstrate the application of XtraGPT in human-AI collaboration. XtraGPT helps the author refine the paper in a controllable manner.
\begin{figure*}[htp]
\centering
\tikzstyle{every node}=[font=\small,scale=0.9]
\begin{tikzpicture}
    \footnotesize
    \definecolor{chatcolor1}{HTML}{FFFACD}
    \definecolor{shadecolor}{gray}{0.95}
    \fontfamily{cmss}\selectfont
    \node[align=left, text width=0.95\textwidth, fill=shadecolor, rounded corners=1mm, draw=black, line width=0.5mm, fill opacity=0.7, text opacity=1, anchor=south east]{
	\textbf{Selected Content:} 
This work proposes HARL, a hierarchical and adaptive reinforcement learning-based auto-scheduler for tensor programs. It is templatefree with improved performance on the generated codes and significantly faster tuning process. The effectiveness of HARL is verified
on different kinds of tensor operators and DNNs, under CPU and GPU hardware platforms. We believe that, HARL is an innovation on how to utilize the advanced reinforcement learning based
decision process to facilitate the time-consuming tensor code optimization.

    \textbf{User Interaction:} 
How can we effectively summarize the key innovations of HARL and their contributions to the field of neural network optimization? What specific results and comparisons to existing methods should we include to demonstrate the significance of our findings?

    };
    \node[align=left, text width=0.95\textwidth, fill=chatcolor1, rounded corners=1mm, draw=black, line width=0.5mm, fill opacity=0.7, text opacity=1, anchor=north east, yshift=-0.5cm,xshift=2cm] {
        \textbf{Responses:}
In this paper, we introduce HARL, a hierarchical and adaptive reinforcement learning-based auto-scheduler that significantly enhances tensor program exploration. By employing a hierarchical RL architecture, HARL enables learning-based decisions at various levels of search granularity, leading to a $22\%$ improvement in tensor operator performance and a $4.3\mathrm{x}$ increase in search speed compared to the state-of-the-art auto-scheduler, Ansor. Our experiments reveal that HARL not only outperforms Ansor in both tuning speed and final program performance but also achieves up to $8\%$ improvement in inference performance on end-to-end neural networks. These results underscore the effectiveness of HARL in addressing the challenges of neural network optimization, providing a robust framework for future research in this domain.
        
        \textbf{Explanations:} 
        The revised conclusion emphasizes the key innovations of HARL, specifically highlighting the $22\%$ performance improvement and $4.3\mathrm{x}$ speed increase over Ansor. By including quantitative results and comparisons to existing methods, we demonstrate the significance of our findings in the context of neural network optimization. The hierarchical RL architecture is clearly linked to these improvements, showcasing how it facilitates more efficient exploration of tensor programs. This alignment with the criteria of effectively summarizing key innovations and providing concrete results is crucial for advancing the field.
        };
\end{tikzpicture}
\caption{A use case on XtraGPT.}
\label{fig:harl}
\end{figure*}

\section{Baseline Model Details}
\label{sec:modeldetails}
Table \ref{tab:baselinedetails} shows the baseline model details.
\begin{table*}[h!]
	\centering

	\resizebox{1\textwidth}{!}{
		\begin{tabular}{lccc}
			\toprule

			\textbf{Models}                                                & \textbf{ Backbone}    & \textbf{GitHub Stars} & \textbf{Huggingface Download} \\
			\midrule
			\multicolumn{4}{c}{\textit{Closed-Source}}                                                                                                     \\
			\midrule
			GPT-4-Turbo~\citep{openai2024gpt4technicalreport}              & \multicolumn{3}{c}{/}                                                         \\
			GPT-4o-mini ~\citep{openai2024gpt4technicalreport}             & \multicolumn{3}{c}{/}                                                         \\
			GPT-3.5-Turbo~\citep{brown2020gpt3}                            & \multicolumn{3}{c}{/}                                                         \\

			\midrule
			\multicolumn{4}{c}{\textit{Open-Source ($>$ 8B)}}                                                                                   \\
			\midrule
			Deepseek-R1 ~\citep{deepseekai2025deepseekr1}                  & Llama                 & 50K                   & 285K                          \\

			Deepseek-V3-671B ~\citep{liu2024deepseekv3}                    & DeepSeek-V3-Base      & 63K                   & 374K                          \\
			Deepseek-V3-32B ~\citep{liu2024deepseekv3}                     & DeepSeek-V3-Base      & 63K                   & 374K                          \\
			Qwen-2-72B-Instruct ~\citep{yang2024qwen2}                     & Qwen-2-72B-Instruct   & 45.3K                 & 374K                          \\
			QwQ-32B-Preview ~\citep{qwq}                                   & Qwen2.5-32B-Instruct  & 15K                   & 198K                             \\
                Phi-4 (14B) ~\citep{abdin2024phi4} & - & - & 557K \\ 

			\midrule
			\multicolumn{4}{c}{\textit{Open-Source ($\leq$8B)}}                                                                                            \\
			\midrule
			Llama-3.1-8B-Instruct ~\citep{grattafiori2024llama3herdmodels} & Llama-3.1-8B-Instruct & 28.1K                 & 5.75M                         \\
			Qwen-2.5-7B-Instruct ~\citep{yang2024qwen2}                    & Qwen-2.5-7B           & 12.6K                 & 1.27M                         \\
			Llama-3.2-3B-Instruct ~\citep{grattafiori2024llama3herdmodels} & Llama-3.2-3B          & 28.1K                 & 1.48M                         \\
			Qwen-2.5-1.5B-Instruct ~\citep{yang2024qwen2}                  & Qwen-2.5-1.5B         & 12.6K                 & 551K                          \\

			\bottomrule
		\end{tabular}}
	\caption{Details information of baseline models. Data collected at 30.1.2025. The "/" indicates that the model uses a private download link or that its download statistics on HuggingFace are not disclosed.}
	\label{tab:baselinedetails}
\end{table*}

\section{Controllability Annotation Criterias and Interface}
\label{sec:ui}

To ensure our data and model quality, We invited three AI experts specializing in inference speedup, graph neural networks (GNN), and Field Programmable Gate Arrays (FPGA) to annotate 5, 3, and 5 papers, respectively. Each paper includes 20 question-answer pairs per model, focusing on section-level improvements. These pairs are distributed across different sections of the paper as follows: 2 for the title, 4 for the abstract, 6 for the introduction, 3 for the background, 3 for the evaluation, and 2 for the conclusion. The controllable criteria used for evaluation are presented in Figure \ref{fig:criterias}.
The annotators' operating interface and the interface of \texttt{XtraGPT} are listed in Figure \ref{fig:ui2}.

\section{Can LLMs Score Full Papers?}
\label{sec:scorepaper}
In the context of academic paper evaluation, the only available human expert review labels at full-paper granularity come from OpenReview. Unfortunately, due to the high cost and inherent biases of human reviews—evidenced by a standard deviation of 1.26 in reviewer ratings for each paper in 2024— it is impractical to invite expert reviewers for every benchmarking scenario that requires full-paper scoring. 

To address this limitation, several studies \citep{lu2024aiscientist,weng2024cycleresearcher,peerreviewmt2024} have explored the use of LLMs for predicting full-paper scores. A key question remains: are LLMs suited to be a reliable reviewer? To investigate this, we follow the approach of \cite{lu2024aiscientist}, applying NeurIPS review guidelines and few-shot examples to assess our test set.

\begin{figure}[h]
    \centering
\begin{minipage}{0.45\columnwidth}
        \centering
        \includegraphics[width=\linewidth]{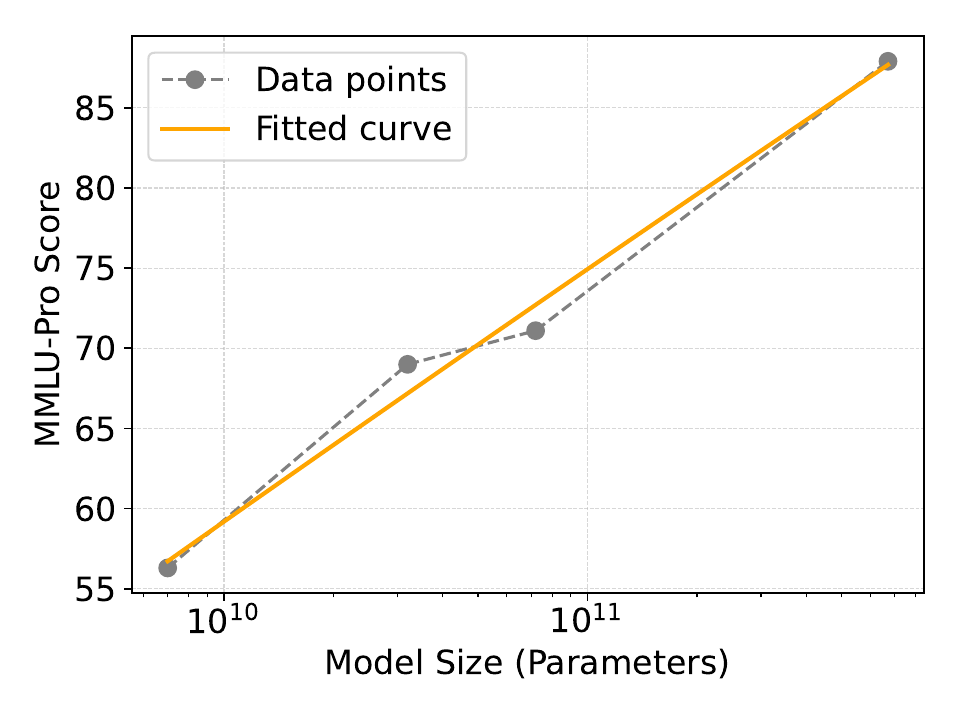}
        \subcaption{MMLU-Pro scores on multi-task understanding across different \# of model parameters}
        \label{fig:llama7b_asqa_train_prob3}
    \end{minipage}
    \hfill
    \begin{minipage}{0.45\columnwidth}
        \centering
        \includegraphics[width=\linewidth]{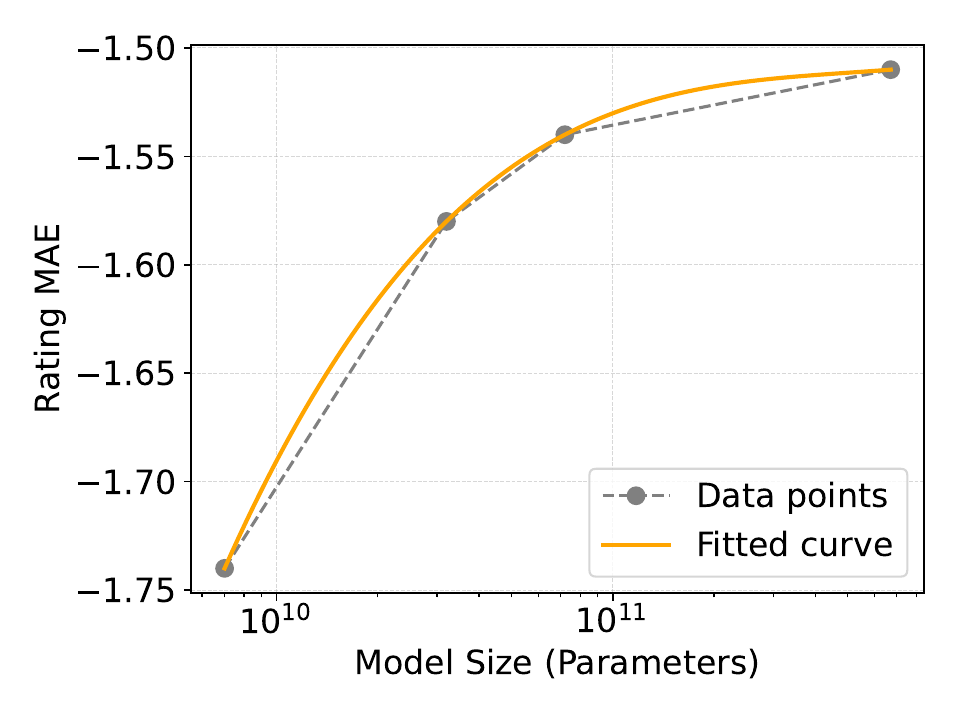}
        \subcaption{MAE scores on paper scoring across different \# of model parameters}
        \label{fig:llama7b_asqa_train_prob4}
    \end{minipage}
    \caption{Scaling trends of Qwen-2.5-7B/32B/72B/Max-Instruct performance. (a) MMLU-Pro scores stably improve with model size. Scaling is effective on multi-task understanding. (b) In the paper scoring task, the rating MAE struggles to go below 1.5. As the model size increases, the reduction in MAE becomes smaller, indicating that scaling offers limited performance improvement.}
    \label{fig:scaling law}
\end{figure}

As shown in Figure \ref{fig:scaling law}, scaling up model parameters is significantly more challenging for paper scoring compared to MMLU-Pro. We can infer that the bottleneck in the paper scoring task \textbf{cannot be simply solved by scaling the model}. LLMs struggle with paper scoring, which is already quite challenging even for human experts (1.16 rating MAE per paper according to \cite{weng2024cycleresearcher}).

\begin{figure}
    \centering
    \includegraphics[width=0.54\textwidth]{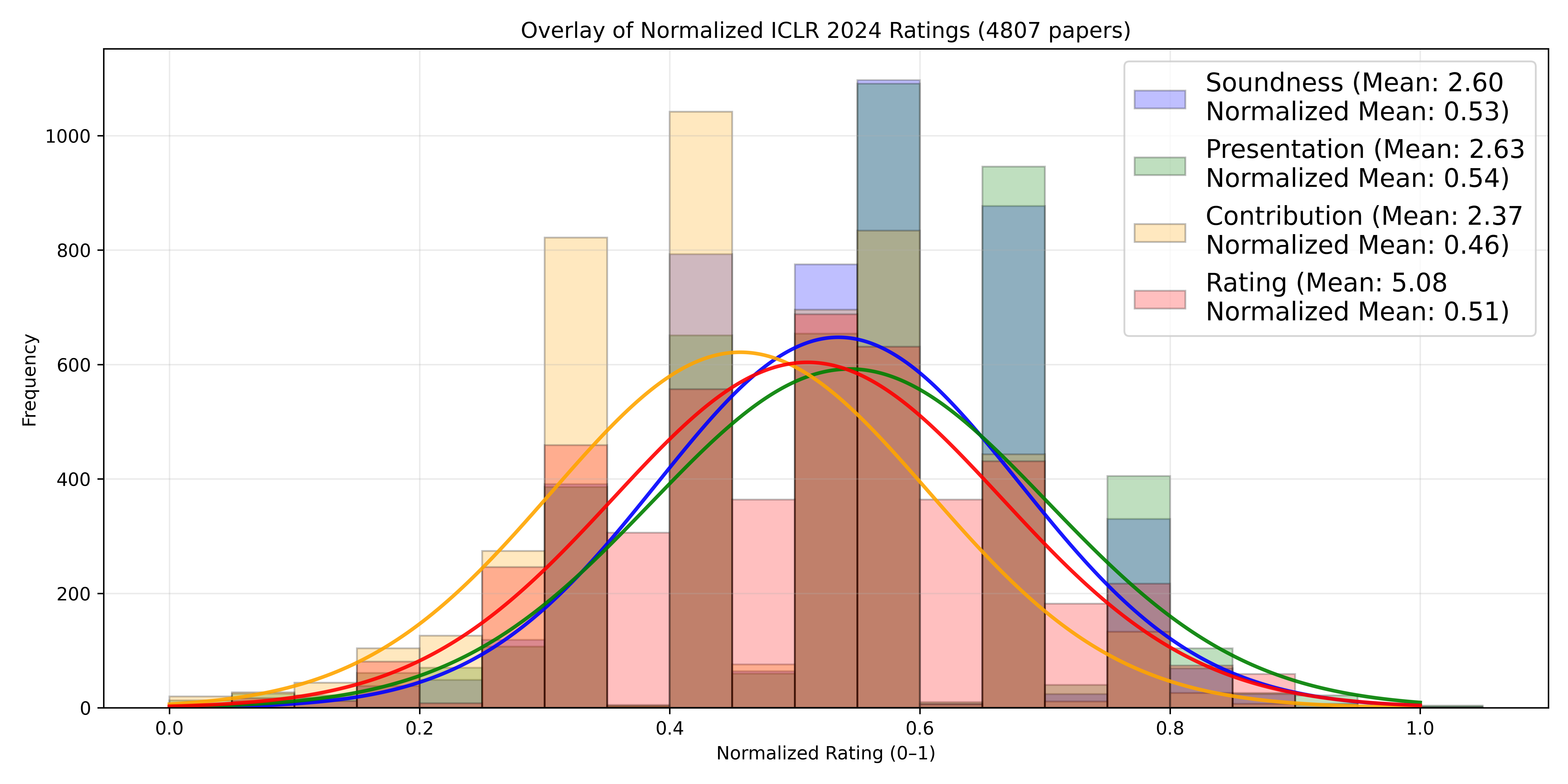}
    \caption{ICLR 2024 paper rating distribution.}
    \label{fig:iclr24distributions}
\end{figure}

\end{document}